\documentclass[]{PHMSociety}

\usepackage{natbib}
\usepackage{apacite}
\usepackage{blindtext}
\usepackage{amsfonts}
\usepackage{subfigure}

\usepackage{fancyhdr}

\usepackage{mathtools}

\DeclarePairedDelimiter\ceil{\lceil}{\rceil}

\begin{document}
\title{Data-driven Prognostics with Predictive Uncertainty Estimation using Ensemble of Deep Ordinal Regression Models}

\author{%
	Vishnu TV, Diksha, Pankaj Malhotra, Lovekesh Vig, and Gautam Shroff
}
\address{
	\affiliation{}{TCS Research, New Delhi, India}{ 
		\email{{\{vishnu.tv, diksha.7,malhotra.pankaj,lovekesh.vig, gautam.shroff\}}@tcs.com}
	}
}

\maketitle
\pagestyle{fancy}
\thispagestyle{plain}

\begin{abstract}%
Prognostics or Remaining Useful Life (RUL) Estimation from multi-sensor time series data is useful to enable condition-based maintenance and ensure high operational availability of equipment.
We propose a novel deep learning based approach for Prognostics with Uncertainty Quantification that is useful in scenarios where: (i)  access to labeled failure data  is scarce due to rarity of failures (ii) future operational conditions are unobserved and (iii) inherent noise is present in the sensor readings. All three scenarios mentioned are unavoidable sources of uncertainty in the RUL estimation process often resulting in unreliable RUL estimates.
To address (i), 
we formulate RUL estimation as an Ordinal Regression (OR) problem, and propose LSTM-OR: deep Long Short Term Memory (LSTM) network based approach to learn the OR function.
We show that LSTM-OR naturally allows for incorporation of censored operational instances in training along with the failed instances, leading to more robust learning.
To address (ii), we propose a simple yet effective approach to quantify predictive uncertainty in the RUL estimation models by training an ensemble of LSTM-OR models. 
Through empirical evaluation on C-MAPSS turbofan engine benchmark datasets, we demonstrate that LSTM-OR is significantly better than the commonly used deep metric regression based approaches for RUL estimation, especially when failed training instances are scarce. 
Further, our uncertainty quantification approach yields high quality predictive uncertainty estimates while also leading to improved RUL estimates compared to single best LSTM-OR models. 
\end{abstract}

\section{Introduction}
In the current digital era, streaming data is ubiquitous. In the context of Industrial Internet of Things, remote health monitoring services driven by sensor driven data analytics are becoming increasingly popular. Data-driven approaches for anomaly detection, diagnostics, prognostics and optimization have been proposed to provide operational support to engineers, ensure high reliability and availability of equipment, and to optimize the operational cost (\cite{da2014internet}).
Typically, a large number of sensors (order of hundreds or sometimes thousands) are installed to capture the operational behavior of complex equipment with various sub-systems interacting with each other.

Recently, deep learning approaches have been proposed for various data-driven health monitoring tasks including anomaly detection (\cite{p:lstm-ad,p:icmlLSTM-AD,gugulothu2018sparse}) and prognostics (\cite{malhotra2016multi,gugulothu2017predicting,zheng2017long}), yielding state-of-the-art results for RUL estimation (\cite{gugulothu2017predicting}) using Recurrent Neural Networks (RNNs).
In this work, we focus on the problem of prognostics or Remaining Useful Life (RUL) estimation 
of operational instances given the current and historical readings from various sensors capturing their behavior.
Deep learning approaches for prognostics, and equipment health monitoring in general, have certain limitations as highlighted in \cite{gugulothu2018on,gugulothu2017predicting,khan2018review}.

In this work, we address two important practical challenges in deep learning based RUL estimation approaches. 
The challenges addressed and the corresponding key contributions of this work are as follows: 

\textbf{Challenge-I}: 
Deep neural networks are prone to overfitting and typically require a large number of labeled training instances to avoid overfitting. 
If failure time for an instance is known, a target RUL can be obtained at any time before the failure time.
However, labeled training instances for RUL estimation are few as failures are rare.
Also, any operational instance (or any instance for which failure time is not known, or which has not failed yet) is considered to be \textit{censored} as target RUL cannot be determined for such an instance.

We note that deep RNNs (\cite{heimes2008recurrent,malhotra2016multi,gugulothu2017predicting,zheng2017long,zhang2018long}) and Convolutional Neural Networks (CNNs) (\cite{babu2016deep}) based approaches formulate RUL estimation as a metric regression (MR) problem where a normalized estimate of RUL is obtained given time series of sensor data via a non-linear regression metric function learned from the data.
This MR formulation of RUL estimation cannot directly leverage censored data typically encountered in RUL estimation scenarios.

\textbf{Key Contribution-I }: In addition to using failed instances for training, we propose a novel approach to \textbf{leverage the censored instances} in a supervised learning setting, in turn, increasing the training data and leading to more robust RUL estimation models. 
We cast RUL estimation as an \textbf{ordinal regression} (\cite{harrell2001ordinal}) problem (instead of the typically used metric regression formulation) and propose LSTM-OR (Long Short Term Memory Networks based Ordinal Regression) based RUL Estimation approach. We show that \textit{partially labeled training instances} can be generated from the readily available operational (non-failed) instances to augment the labeled training data in the ordinal regression setting to build more robust RUL estimation models. We empirically show that LSTM-OR outperforms LSTM-MR by effectively leveraging censored data when the number of failed instances available for training is small. 

\textbf{Challenge-II}: 
The black-box nature of deep neural networks makes it difficult to interpret the predictions/estimates \cite{}, and in turn, gauge the reliability of the predictions. 
It is, therefore, desirable to \textbf{quantify the predictive uncertainty} in deep neural network based predictions of RUL - it can aid engineers and operators in risk assessment and decision making while accounting for the reliability of predictions. 

\textbf{Key Contribution-II}: We propose a simple yet effective approach to \textbf{quantify uncertainty based on an ensemble of LSTM-OR models} (using similar idea as in \cite{NIPS2017_7219} as detailed in Section \ref{sec:uncertaintyQunatification}). Ensemble of deep LSTM-OR models leads to improved RUL estimation performance, and also the empirical standard deviation (ESD) of the predictions from LSTM-OR models provides an approximate measure of uncertainty. 
We empirically show that when ESD (i.e. the uncertainty in estimation) is low, the corresponding error in estimation is also low; making ESD a useful uncertainty quantification metric. 

\textbf{Organization of the paper}: We provide an overview of related literature in Section \ref{sec:rw}.
In Section \ref{sec:lstm}, we briefly introduce deep LSTM networks as used to build our deep OR models.
We provide details of LSTM-OR and uncertainty quantification approaches in Sections \ref{sec:deepOR} and \ref{sec:uncertaintyQunatification}, respectively. We provide experimental evaluation details and observations in Section \ref{sec:exp}, and finally conclude in Section \ref{sec:conc}.

\section{Related Work\label{sec:rw}}

\textit{Trajectory Similarity based RUL estimation}: An important class of approaches for RUL estimation is based on trajectory similarity, e.g. \cite{wang2008similarity,khelif2014rul,lam2014enhanced,malhotra2016multi,gugulothu2017predicting}. 
These approaches compare the health index trajectory or trend of a test instance with the trajectories of failed train instances to estimate RUL using a distance metric such as Euclidean distance.
Such approaches work well when trajectories are smooth and monotonic in nature but are likely to fail in scenarios when there is noise or intermittent disturbances (e.g. spikes, operating mode change, etc.) as the distance metric may not be robust to such scenarios (\cite{gugulothu2017predicting}).

\textit{Metric Regression based RUL estimation}: Another class of approaches is based on metric regression.
Unlike trajectory similarity based methods which rely on comparison of trends, metric regression methods attempt to learn a function to directly map sensor data to RUL, e.g. \cite{heimes2008recurrent,benkedjouh2013remaining,dong2014lithium,babu2016deep,gugulothu2017predicting,zheng2017long,vishnu2018recurrent}.
Such methods can better deal with non-monotonic and noisy scenarios by learning to focus on the relevant underlying trends irrespective of noise. 
Within metric regression methods, few methods consider non-temporal models such as Support Vector Regression for learning the mapping from values of sensors at a given time instance to RUL, e.g. \cite{benkedjouh2013remaining,dong2014lithium}.

\textit{Temporal models for RUL estimation}: Deep temporal models such as those based on RNNs (\cite{heimes2008recurrent,malhotra2016multi,gugulothu2017predicting,zheng2017long}) or Convolutional Neural Networks (CNNs) (\cite{babu2016deep}) can capture the degradation trends better compared to non-temporal models, and are proven to perform better.
Moreover, these models can be trained in an end-to-end learning manner without requiring feature engineering.
Despite all these advantages of deep models, they are prone to overfitting in often-encountered practical scenarios where the number of failed instances is small, and most of the data is censored. 
Our approach based on ordinal regression provisions for dealing with such scenarios, by using censored instances in addition to failed instances to obtain more robust models.

\textit{Ordinal Regression for Survival Analysis}: Ordinal Regression has been extensively used for applications such as age estimation from facial images (\cite{chang2011ordinal,yang2013automatic,niu2016ordinal,liu2017ordinal}), however the applications are restricted to non-temporal image data using Convolutional Neural Networks.
\cite{cheng2008neural,luck2017deep} use feed-forward neural networks based ordinal regression for survival analysis.
To the best of our knowledge, the proposed LSTM-OR approach is the first attempt to leverage ordinal regression based training using temporal LSTM networks for RUL estimation.

\textit{Deep Survival Analysis}: A set of techniques for deep survival analysis have been proposed in the medical domain, e.g. \cite{katzman2018deepsurv,luck2017deep}.
On similar lines, an approach to combine deep learning and survival analysis for asset health management has been proposed in \cite{liao2016combining}.
However, it is not clear as to how such approaches can be adapted for RUL estimation applications, as they focus on estimating the survival probability at a given point in time, and cannot provide RUL estimates.
Further, \cite{chapfuwa2018adversarial} proposes an approach that leverages adversarial learning for doing time-event modeling in health domain.   
On the other hand, LSTM-OR is capable of providing RUL estimates using time series sensor data.

\textit{Uncertainty quantification in RUL estimation models}: Uncertainty analysis in data-driven equipment health monitoring is an active area of research and an unsolved problem. The approaches described in \cite{sankararaman2013novel}, \cite{6496971} use analytical algorithms, unlike sampling-based methods, to estimate the uncertainty in prognostics. They consider various sources of uncertainty such as the loading and operating conditions of the system at hand, inaccurate sensor measurements, etc. to quantify their combined effect on RUL predictions.
The task is formulated as an uncertainty propagation problem where the various types of uncertainty are propagated through state space models until failure. Also, the future states of the system are estimated using the state space models and are used to arrive at an estimate of RUL.
Unlike these approaches, we focus on estimating RUL as well as predictive uncertainty by using an ensemble of deep neural networks to model the time-series of sensor data available till a given point in time, without predicting the future states of the system.
Our approach does not rely on any assumptions such as those needed in a state-space model.
Further, domain knowledge of the underlying dynamics of a system is not needed to quantify uncertainty, and therefore, our approach is much simpler to adapt.

\textit{Uncertainty quantification for deep neural networks}: Recently, \cite{gal2016dropout} proposed the use of dropout at the inference time to provide Bayesian approximation in the RUL estimation.  
Further, \cite{NIPS2017_7219} proposed the use of an ensemble of neural networks for predictive uncertainty estimation and demonstrated their use in comparison to Bayesian methods. Similarly, we also use an ensemble of LSTM networks to estimate the empirical uncertainty in RUL predictions.

\section{Background: Deep LSTM Networks\label{sec:lstm}}
We use a variant of LSTMs (\cite{hochreiter1997long}) as described in \cite{zaremba2014recurrent} in the hidden layers of the neural network.
Hereafter, we denote column vectors by bold small letters and matrices by bold capital letters.
For a hidden layer with $h$ LSTM units, the values for the input gate $\mathbf{i}_t$, forget gate $\mathbf{f}_t$, output gate $\mathbf{o}_t$, hidden state $\mathbf{z}_t$, and cell state $\mathbf{c}_t$ at time $t$ are computed using the current input $\mathbf{x}_t$, the previous hidden state $\mathbf{z}_{t-1}$, and the cell state $\mathbf{c}_{t-1}$, where $\mathbf{i}_t$, $\mathbf{f}_t$, $\mathbf{o}_t$, $\mathbf{z}_t$, and $\mathbf{c}_t$ are real-valued $h$-dimensional vectors.

Consider $W_{n_1,n_2}:\mathbb{R}^{n_1} \rightarrow \mathbb{R}^{n_2}$ to be an affine transform of the form $\mathbf{z}\mapsto \mathbf{Wz}+\mathbf{b}$ for matrix $\mathbf{W}$ and vector $\mathbf{b}$ of appropriate dimensions.
In the case of a multi-layered LSTM network with $L$ layers and $h$ units in each layer, the hidden state $\mathbf{z}_{t}^{l}$ at time $t$ for the $l$-th hidden layer is obtained from the hidden state at $t-1$ for that layer $\mathbf{z}_{t-1}^{l}$ and the hidden state at $t$ for the previous ($l-1$)-th hidden layer $\mathbf{z}_{t}^{l-1}$. 
The time series goes through the following transformations iteratively at $l$-th hidden layer for $t=1$ through $T$, where $T$ is length of the time series:

\begin{equation}\label{eq:lstm1}
  \left(\begin{aligned}
   \mathbf{i}_t^l\\
   \mathbf{f}_t^l\\
   \mathbf{o}_t^l\\
   \mathbf{g}_t^l
  \end{aligned}\right)=\left(\begin{aligned}
   \sigma\quad\\
   \sigma\quad\\
   \sigma\quad\\
   tanh\\
  \end{aligned}\right)W_{2h,4h}
  \left(\begin{aligned}
    \mathbf{D(z}_{t}^{l-1})\\
    \mathbf{z}_{t-1}^l\\
    \end{aligned}\right)
\end{equation}
where the cell state $\mathbf{c}_t^l$ is given by $\mathbf{c}_t^l=\mathbf{f}_t^l\mathbf{c}_{t-1}^l+\mathbf{i}_t^l\mathbf{g}_t^l$, and the hidden state $\mathbf{z}_t^l$ is given by $\mathbf{z}_t^l=\mathbf{o}_{t}^ltanh(\mathbf{c}_t^l)$.
We use dropout for regularization (\cite{pham2014dropout}), which is applied only to the non-recurrent connections, ensuring information flow across time-steps for any LSTM unit. 
The dropout operator $\mathbf{D}(\cdot)$ randomly sets the dimensions of its argument to zero with probability equal to a dropout rate.
The sigmoid ($\sigma$) and $tanh$ activation functions are applied element-wise. 

In a nutshell, this series of transformations for $t=1\ldots T$, converts the input time series $\mathbf{x}=\mathbf{x}_1\ldots\mathbf{x}_T$ of length $T$ to a fixed-dimensional vector $\mathbf{z}_T^L \in \mathbb{R}^h$.
We, therefore, represent the LSTM network by a function $f_{LSTM}$ such that $\mathbf{z}_T^L = f_{LSTM}(\mathbf{x};\mathbf{W})$, where $\mathbf{W}$ represents all the parameters of the LSTM network.

\begin{figure*}[h]
 \subfigure[Metric Regression]{\includegraphics[trim={25cm 3cm 2cm 2cm},clip,width=0.3\textwidth]{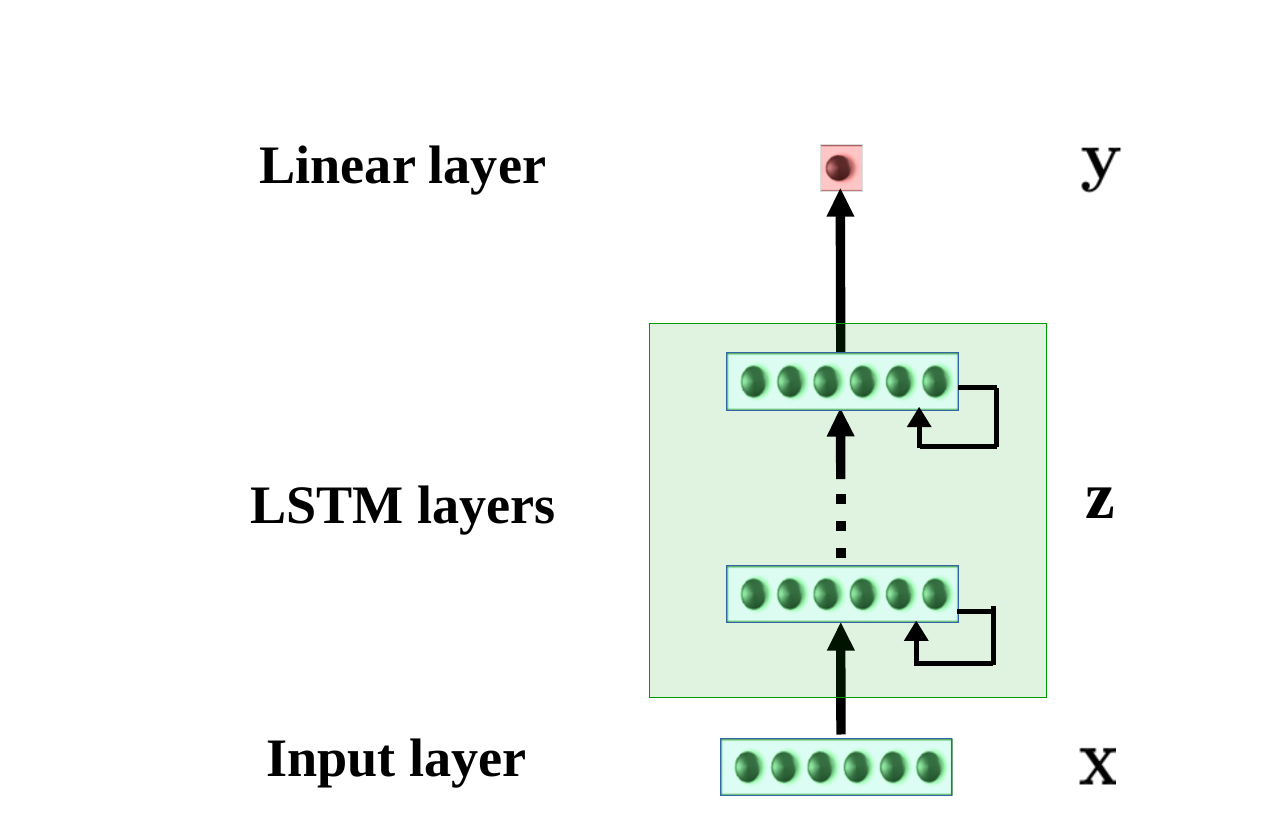}}
 \subfigure[Ordinal Regression]{\includegraphics[trim={2cm 3cm 25cm 0cm},clip,width=0.3\textwidth]{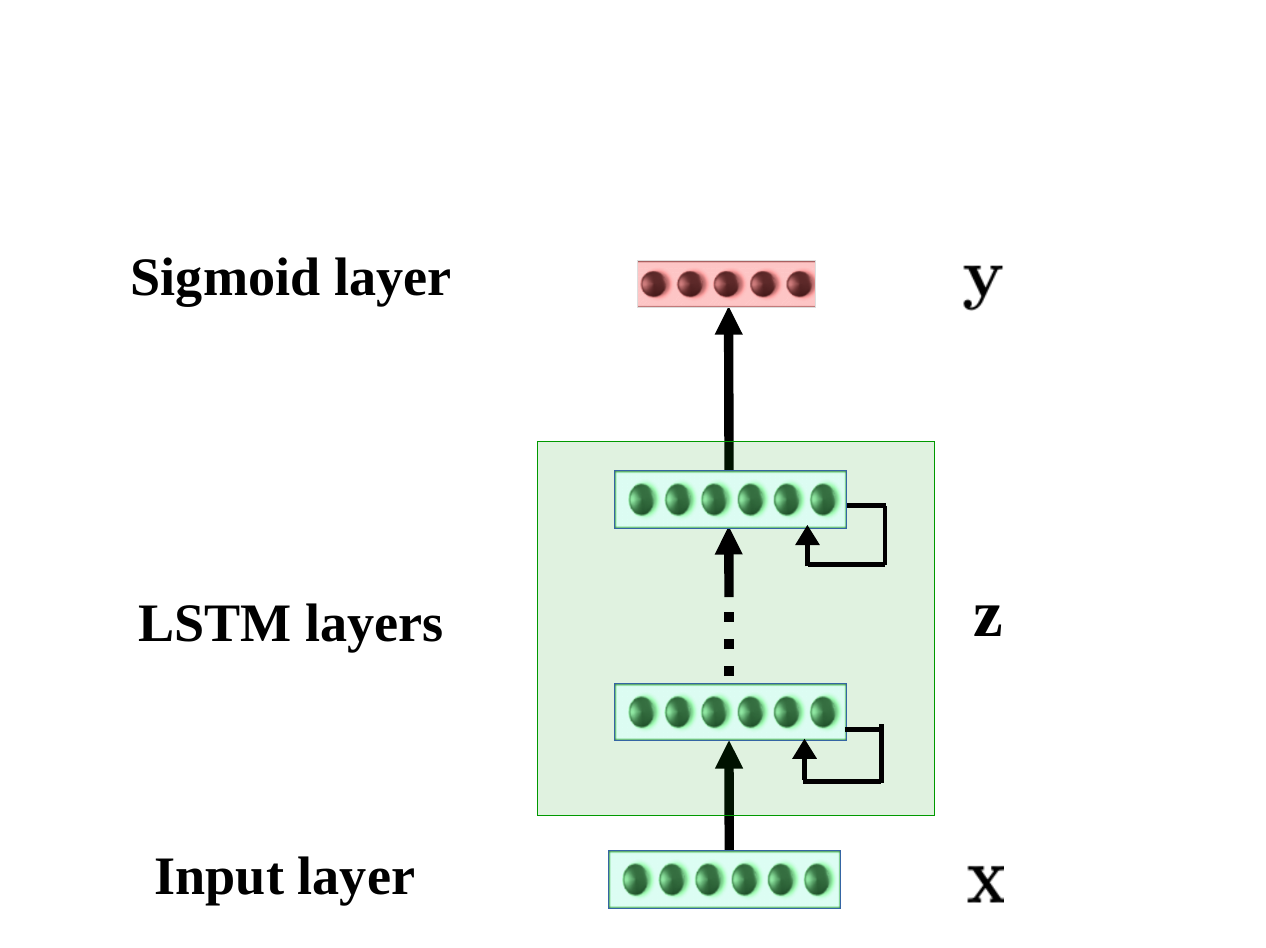}}
 \subfigure[Ordinal Regression For Censored Data]{\includegraphics[trim={2cm 3cm 25cm 0cm},clip,width=0.3\textwidth]{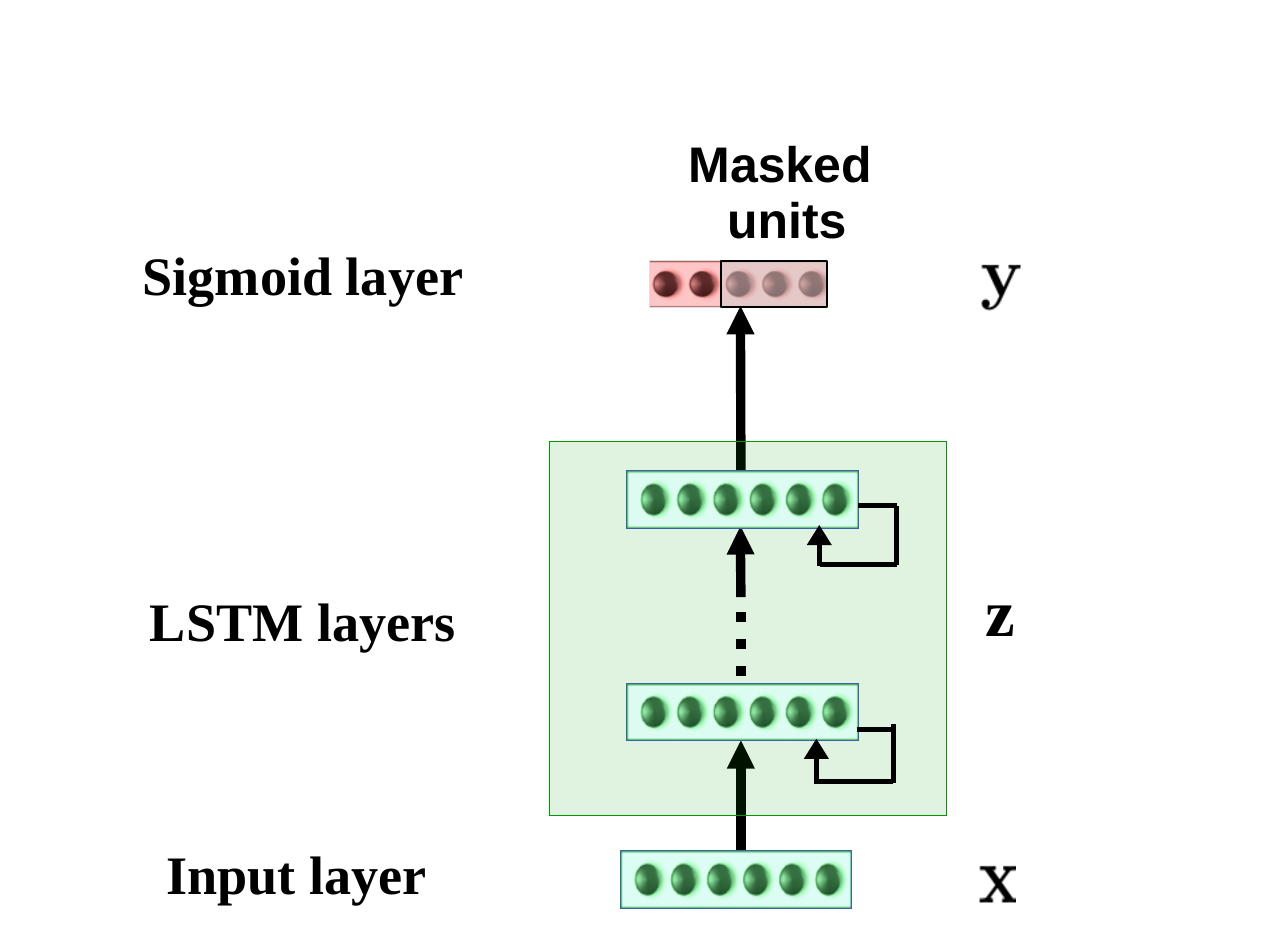}} 
 \caption{Deep Ordinal Regression versus Deep Metric Regression.}
\end{figure*}

\begin{figure*}[h]
\subfigure[Process overview for LSTM-OR.\label{fig:flowchart}]{\includegraphics[trim={0cm 1.5cm 0cm 1cm},clip,width=0.9\columnwidth]{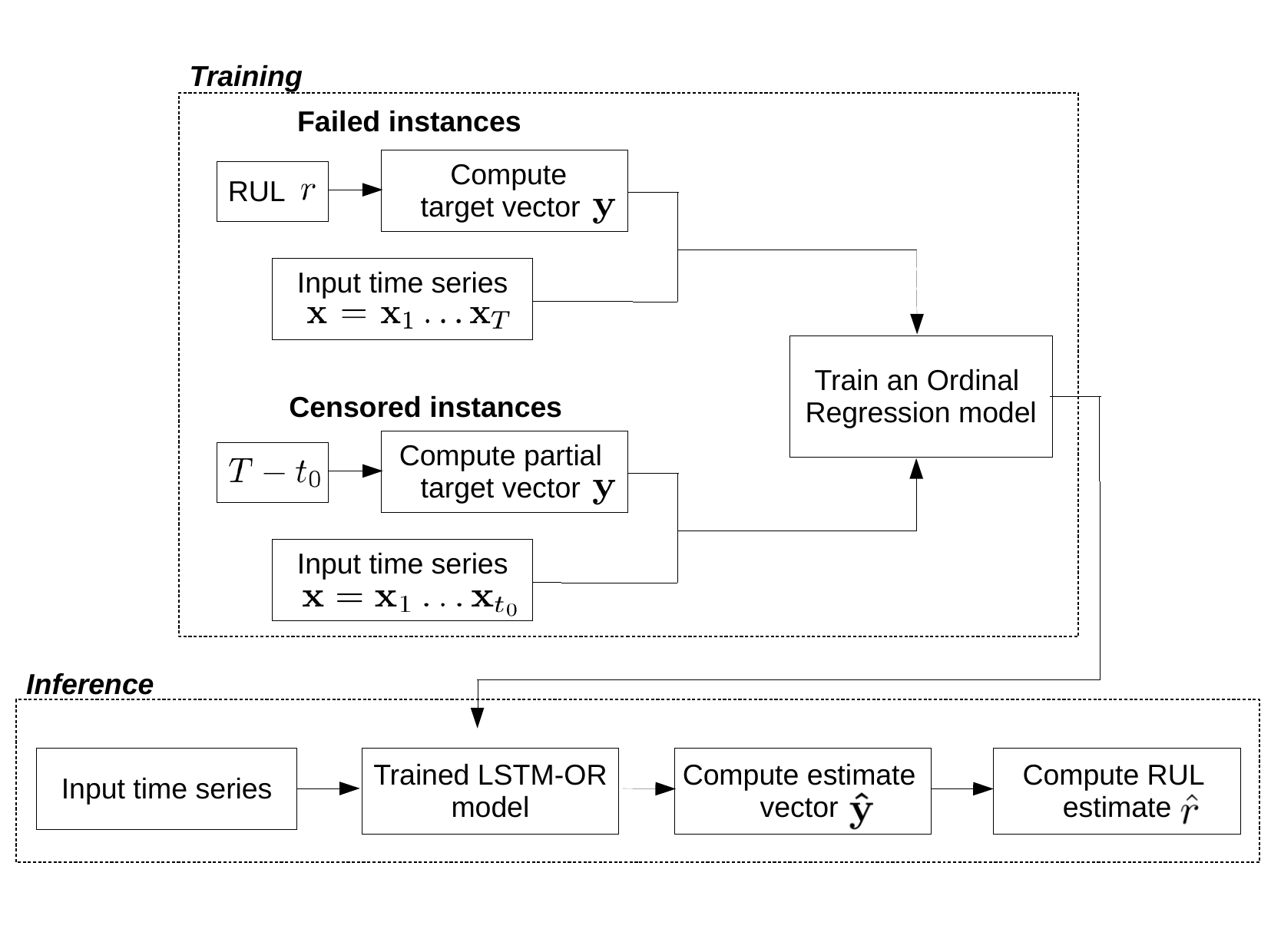}}
\subfigure[RUL and Uncertainty Estimation using Ensemble of LSTM-OR models.\label{fig:flowchart-orce}]{\includegraphics[trim={0cm 1.5cm 0cm 1cm},clip,width=1.1\columnwidth]{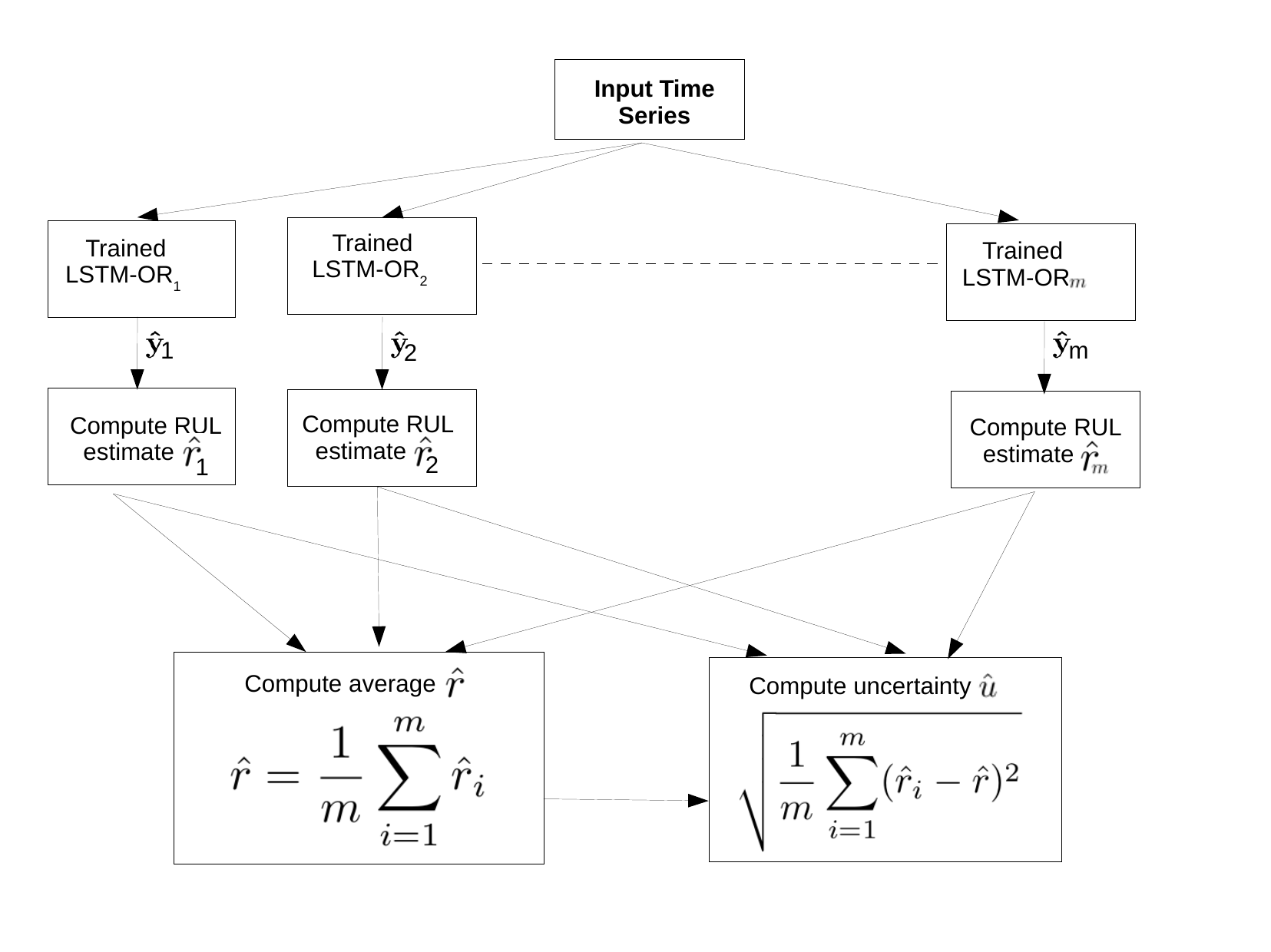}}
\caption{Steps in LSTM-OR and Ensemble of LSTM-OR.}
\end{figure*}

\section{Deep Ordinal Regression for RUL Estimation\label{sec:deepOR}}
\subsection{Terminology}
Consider a learning set $\mathcal{D} = \{\mathbf{x}^{i},r^{i}\}^n_{i=1}$ of $n$ failed instances, where $r^i$ is the target RUL, $\mathbf{x}^{i}= \mathbf{x}^i_{1} \ldots \mathbf{x}^i_{{T^i}} \in \mathcal{X}$ is a multivariate time series of length $T^i$, $\mathbf{x}^i_{t} \in \mathbb{R}^{p}$, $p$ is the number of input features (sensors). 
The total operational life of an instance $i$ till the failure point is $F^i$, s.t. $T^i \leq F^i$. 
Therefore, $r^{i}=F^i-T^i$ is the RUL in given unit of measurement, e.g., number of cycles or operational hours.
Hereafter, we omit the superscript $i$ in this section for better readability, and provide all the formulation considering an instance (unless stated otherwise).

We consider an upper bound $r_u$ on the possible values of RUL as, in practice, it is not possible to predict too far ahead in future. So if $r > r_u$, we clip the value of $r$ to $r_u$.
The usually defined goal of RUL estimation via Metric Regression (MR) is to learn a mapping $f_{MR}: \mathcal{X} \rightarrow [0,r_u]$.
With these definitions, we next describe LSTM-based Ordinal Regression (LSTM-OR) approach as summarized in Figure \ref{fig:flowchart}, and then describe how we incorporate censored data into the LSTM-OR formulation.

\subsection{LSTM-based Ordinal Regression}

\begin{figure}[h]
\subfigure[Failed Instance\label{fig:illus_failed}]{\includegraphics[trim={12cm 2cm 2cm 2cm},clip,width=0.45\columnwidth]{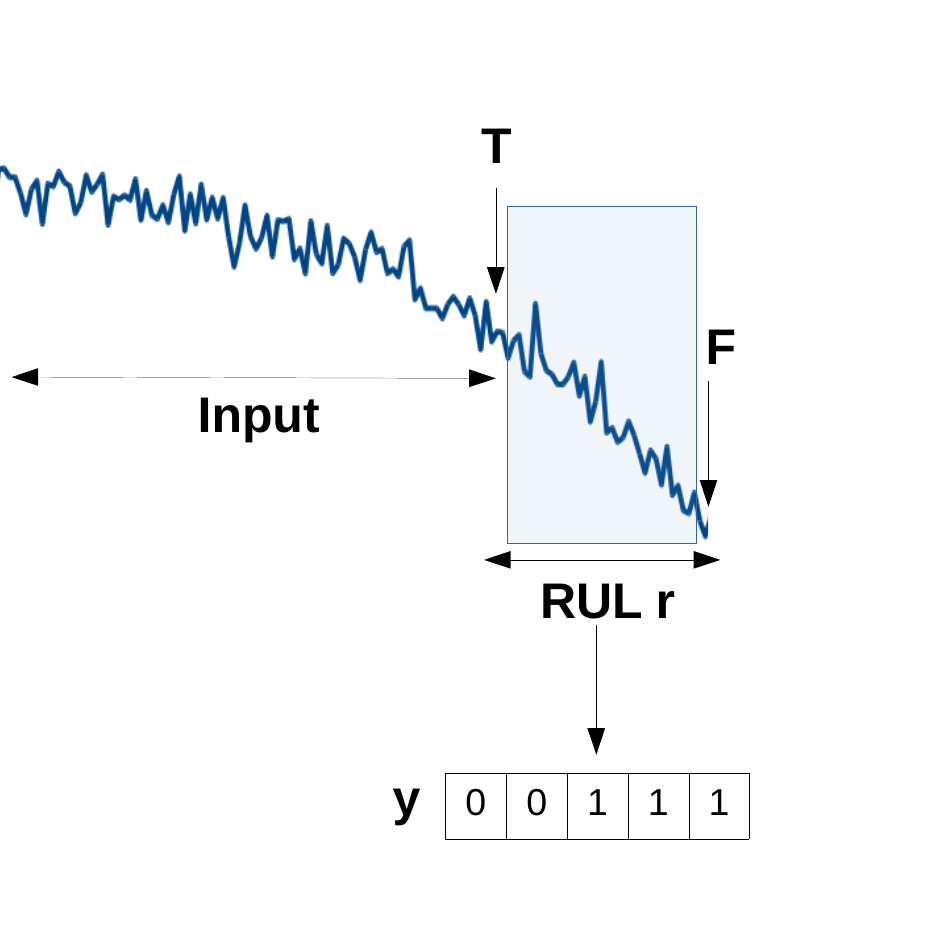}}
 \subfigure[Censored Instance\label{fig:illus_cens}]{\includegraphics[trim={12cm 2.2cm 2cm 2cm},clip,width=0.45\columnwidth]{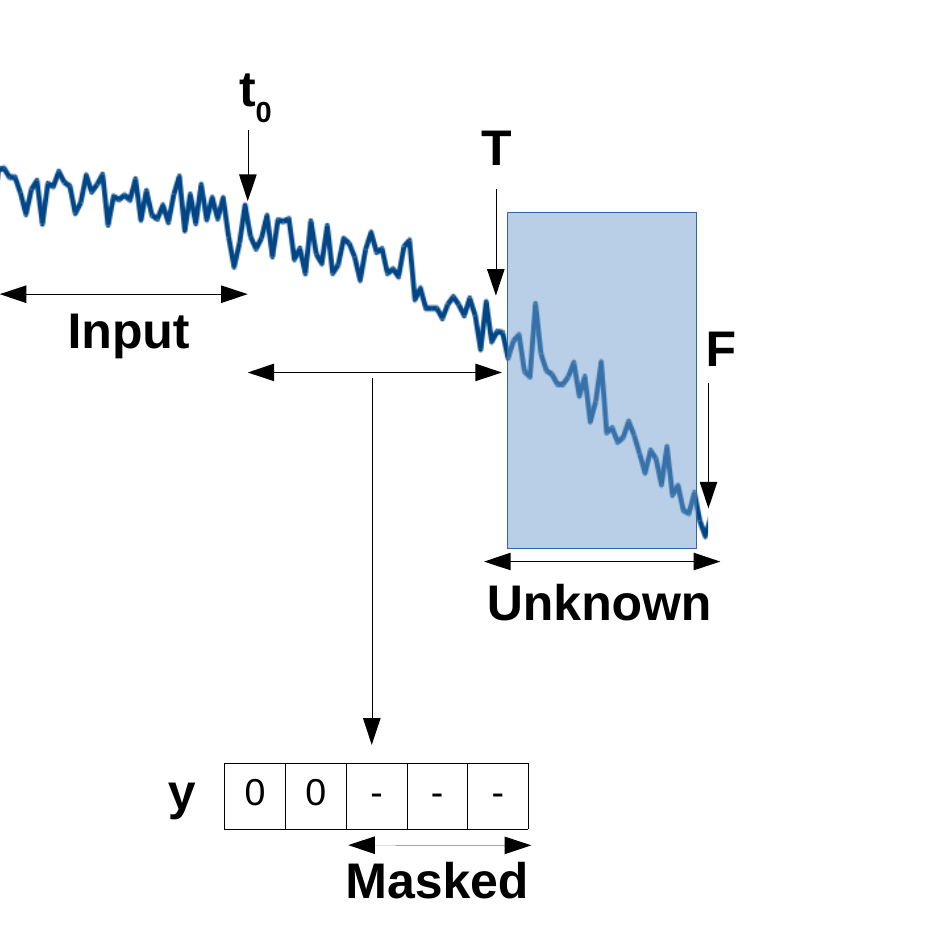}}
 \caption{Target vector creation for failed versus censored instance.\label{fig:illus}}
\end{figure}

Instead of mapping an input time series to a real-valued number as in MR, we break the range $[0,r_u]$ of RUL values into $K$ intervals of length $c$ each, where each interval is then considered as a discrete variable.
The $j$-th interval corresponds to $((j-1)\frac{r_u}{c},j\frac{r_u}{c}]$, and $r$ is mapped to the $k$-th interval with $k=\ceil*{\frac{r}{c}}$, where $\ceil*{.}$ denotes the ceiling function.

We consider $K$ binary classification sub-problems for the $K$ discrete variables (intervals): a classifier $C_j$ solves the binary classification problem of determining whether $r\leq j\frac{r_u}{c}$.

We train an LSTM network for the $K$ binary classification tasks simultaneously by modeling them together as a multi-label classification problem:
We obtain the multi-label target vector $\mathbf{y}=[y_{1},\ldots,y_{K}] \in \{0,1\}^K$  from $r$ such that
\begin{equation}\label{eq:target}
  y_{j} =
    \begin{cases}
      0 & j<k\\
      1 & j\geq k\\
    \end{cases}       
\end{equation}

where $j=1,2, \dots,K$.

For example, consider a scenario where $K=5$, and 
$r$ maps to the third interval such that $k=3$. 
The target is then given by $\mathbf{y}=[0,0,1,1,1]$, as illustrated in Figure \ref{fig:illus_failed}.
Effectively, the goal of LSTM-OR is to learn a mapping $f_{OR}: \mathcal{X} \rightarrow \{0,1\}^K$
by minimizing the loss function $\mathcal{L}_{OR}$ given by:

\begin{equation}\label{eq:OR}
\begin{aligned}
\mathbf{z}_{T}^L&=f_{LSTM}(\mathbf{x};\mathbf{W})\\
\mathbf{\hat{y}}&= \sigma(\mathbf{W}_C\:\mathbf{z}_{{T}}^L+\mathbf{b}_C)\\
\mathcal{L}_{OR}(\mathbf{y},\mathbf{\hat{y}})&=-\frac{1}{K}\sum_{j=1}^{K}y_{j}\cdot log(\hat{y}_{j})+(1-y_{j})\cdot log(1-\hat{y}_{j})\\
\end{aligned}
\end{equation}
where, $\mathbf{\hat{y}}$ is the estimate for target $\mathbf{y}$, $\mathbf{W}$ represents the parameters of the LSTM network, and $\mathbf{W}_C$ and $\mathbf{b}_C$ are the parameters of the layer that maps $\mathbf{z}_{T}^L$ to the output sigmoid layer.

\subsection{Using Censored Data for Training}
For any censored instance, the data is available only till a time $T$ prior to failure and the failure time $F$ is unknown (illustrated in Figure \ref{fig:illus_cens}).
Therefore, the target RUL $r$ is also unknown. 
However, at any time $t_0$ s.t. $1\leq t_0<T$, it is known that the RUL $r > T-t_0$ since the instance is operational at least till $T$.
Considering $\mathbf{x}=\mathbf{x}_{1} \ldots \mathbf{x}_{t_0}$ as the input time series, we next show how we assign labels to few of the dimensions $y_j$ of the target vector $\mathbf{y}$:
Assuming $T-t_0$ maps to the interval $k'=\ceil*{\frac{T-t_0}{c}}$, 
since $T-t_0 < r$, we have $\ceil*{\frac{T-t_0}{c}} \leq \ceil*{\frac{r}{c}} \implies k^{\prime} \leq k$.
Since $k$ is unknown (as $r$ is unknown) and we have $k^{\prime} \leq k$, the target vector $\mathbf{y}$ can only be partially obtained: 

\begin{equation}\label{eq:maskedtarget}
  y_{j} =
    \begin{cases}
      0 & j<k'\\
      unknown & j\geq k'\\
    \end{cases}       
\end{equation}

For all $j \geq k^{\prime}$, the corresponding binary classifier targets are masked, as shown in Figure \ref{fig:illus_cens}, and the outputs from these classifiers are not included in the loss function for the instance. 
The loss function $\mathcal{L}_{ORC}$ given by Equation \ref{eq:OR} can thus be modified for including the censored instances in training as:
\begin{equation}\label{eq:maskedL}
 \mathcal{L}_{ORC}(\mathbf{y},\mathbf{\hat{y}})=-\frac{1}{K'}\sum_{j=1}^{K'}y_{j}\cdot log(\hat{y}_{j})+(1-y_{j})\cdot log(1-\hat{y}_{j})\\
\end{equation}
where $K'=k'-1$ for a censored instance and $K'=K$ for a failed instance.

\subsection{Mapping OR estimates to RUL}
Once trained, each of the $K$ classifiers provides a probability $\hat{y}_j$ for RUL being greater than the upper limit of the interval corresponding to the $j$-th classifier.
We obtain the point-estimate $\hat{r}$ for ${r}$ from $\hat{\mathbf{y}}$ for a test instance as follows (similar to \cite{chang2011ordinal}): 
\begin{equation}\label{eq:pointEst}
\hat{r} = r_u(1-\frac{1}{K}\sum_{j=1}^{K}\hat{y}_{j}) 
\end{equation}
It is worth noting that once learned, the LSTM-OR model can be used in an online manner for operational instances: at current time instance $t$, the sensor data from the latest $T$ time instances can be input to the model to obtain the RUL estimate $r$ at $t$.

\section{Predictive Uncertainty Quantification using Ensemble of LSTM-OR Models \label{sec:uncertaintyQunatification}}
Uncertainty quantification is very important in case of RUL estimation as equipment and operations involved are often of critical nature, and reliable predictions close to (but of course, prior to) failures can help avoid catastrophic failures by generating suitable alarms beforehand. 
Lack of sufficient training data, inherent noise in sensor readings, and uncertainty in the future usage and operation of equipment are few sources of uncertainty in case of data-driven predictive models for RUL estimation. 
Quantifying uncertainty in RUL estimates can assist ground engineers and operators to arrive at more informed decisions compared to scenarios where only RUL estimates are available without any metric indicating whether the model is certain about the estimate or not. 
In other words, uncertainty quantification of the RUL estimate enhances the reliability of data-driven models. 
This is even more relevant in deep neural network-based estimation models due to their otherwise black-box nature.

An uncertainty metric can be considered to be reliable if: i) for low uncertainty values, i.e. whenever the model is confident about its estimations, the corresponding errors in the RUL estimations are low, and for high uncertainty values, the corresponding errors in the RUL estimation model should be high, ii) it produces RUL estimates with low uncertainty when a failure is approaching, i.e. the model should be able to precisely estimate the RUL with a high degree of certainty close to failures.

To quantify the predictive uncertainty in the target vector estimate $\hat{\mathbf{y}}$ and the corresponding RUL estimate $\hat{r}$, we consider training an ensemble of LSTM-OR models.
We consider an ensemble learning approach similar to that introduced in \cite{NIPS2017_7219}:
For training an ensemble of LSTM-OR models, we consider all the training data while using different (random) initializations of the parameters ($\mathbf{W},\mathbf{W}_C,\mathbf{b}_C$) of LSTM-OR models and random shuffling of the training instances to obtain $m$ different models in an ensemble. 
The final RUL estimate of the ensemble is given by simple average of the RUL estimates of the $m$ models in the ensemble, and the empirical standard deviation (ESD) in the RUL estimates is used as an approximation of the predictive uncertainty in RUL estimation.
More specifically, as shown in Figure \ref{fig:flowchart-orce}, we train $m$ LSTM-OR models such that we have $m$ RUL estimates $\hat{r_{i}}$ for any instance, $i=1,\ldots,m$.
We obtain the point estimate $\hat{r}$ for ${r}$ from $\hat{r_{i}}$ for an instance as follows: 

\begin{equation}\label{eq:uncertaintyEst}
\hat{r} = \frac{1}{m}\sum_{i=1}^{m}\hat{r}_{i}
\end{equation}

The uncertainty $\hat{u}$ in terms of ESD is given by:
\begin{equation}\label{eq:uncertaintyEstStdDev}
\hat{u}_{ESD} = \sqrt{\frac{1}{m}\sum_{i=1}^{m}(\hat{r}_{i} - \hat{r})^2}
\end{equation}
We normalize the uncertainty values ($\hat{u}_{ESD}$) using the minimum and maximum uncertainty values across all instances in a hold-out validation set through min-max normalization.
We also consider other measures of uncertainty quantification in terms of entropy (similar to \cite{park2015using}) as explained in  Appendix \ref{apx:ent_uncertainty} but found ESD to be the most robust measure of uncertainty. We support this with experimental evaluation in Section \ref{sec:uncertainty}.

\section{Experimental Evaluation\label{sec:exp}} 
We evaluate RUL estimation and uncertainty quantification approaches using the publicly available C-MAPSS aircraft turbofan engine benchmark datasets (\cite{saxena2008turbofan}).
We provide an overview of the dataset in Section \ref{sec:dd}.
We consider metric regression models and ordinal regression models trained only on failed instances as baseline models, and compare following approaches for RUL estimation: i) \textbf{MR}: LSTM-MR using failed instances only (as in \cite{zheng2017long,heimes2008recurrent,gugulothu2017predicting}), ii) \textbf{OR}: LSTM-OR using failed instances only and using loss as in Equation \ref{eq:OR}, iii) \textbf{ORC}: LSTM-OR leveraging censored data along with failed instances using loss as in Equation \ref{eq:maskedL}, iv) \textbf{ORCE}: simple average ensemble of ORC models.
We describe RUL estimation approaches in Section \ref{sec:RUL}.
Further, to evaluate uncertainty quantification approach as described in Section \ref{sec:uncertaintyQunatification}, we study the relationship of uncertainty estimates with error and ground truth RUL in Section \ref{sec:uncertainty} while also introducing novel metrics to evaluate the efficacy of uncertainty estimates in context of prognostics.

\subsection{Dataset Description\label{sec:dd}}
We consider datasets FD001 and FD004 from the simulated turbofan engine datasets\footnote{\url{https://ti.arc.nasa.gov/tech/dash/groups/pcoe/\\prognostic-data-repository/\#turbofan}} (\cite{saxena2008turbofan}). 
The training sets (\textit{train\_FD001.txt} and \textit{train\_FD004.txt}) of the two datasets contain time series of readings for 24 sensors (21 sensors and 3 operating condition variables) of several instances (100 in FD001 and 249 in FD004) of a turbofan engine from the beginning of usage till end of life.  
The time series for the instances in the test sets (\textit{test\_FD001.txt} and \textit{test\_FD004.txt}) are pruned some time prior to failure, such that the instances are operational and their RUL needs to be estimated. 
The actual RUL values for the test instances are available in \textit{RUL\_FD001.txt} and \textit{RUL\_FD004.txt}.
We randomly sample 20\% of the available training set instances, as given in Table \ref{tab:datastats}, to create a validation set for hyperparameter selection.

For simulating the scenario for censored instances, a percentage $p_c \in \{0,50,70,90\}$ of the training and validation instances are randomly chosen, and time series for each instance is randomly truncated at one point prior to failure. 
We then consider these truncated instances as censored (currently operational) and their actual RUL values as unknown. 
The remaining $(100-p_c)$\% of the instances are considered as failed.
Further, the time series of each instance thus obtained (censored and failed) is truncated at 20 random points in the life prior to failure, and the exact RUL $r$ for failed instances and the minimum possible RUL $T-t_0$ for the censored instances (as in Section \ref{sec:deepOR} and Figure \ref{fig:illus}) at the truncated points are used for obtaining the models. 
The number of instances thus obtained for training and validation for $p_c=0$ is given in Table \ref{tab:winstats}.
The test set remains the same as the benchmark dataset across all scenarios (with no censored instances).
The MR and OR approaches cannot utilize the censored instances as the exact RUL targets are unknown, while ORC can utilize the lower bound on RUL targets to obtain partial labels as per Equation \ref{eq:maskedtarget}.

An engine may operate in different operating conditions and also have different failure modes at the end of its life. The number of operating conditions and failure modes for both the datasets are given in Table \ref{tab:datastats}.
FD001 has only one operating condition, so we ignore the corresponding three sensors such that $p=21$, whereas FD004 has six operating conditions determined by the three operating condition variables. We map these six operating conditions to a 6-dimensional one hot vector as in \cite{zheng2017long}, such that $p=27$.

\subsection{RUL Estimation \label{sec:RUL}}
In this section, we define performance metrics to evaluate our RUL estimation models i.e ORC and ORCE. Further, we discuss our experimental settings which is followed by results and observations. We also draw a comparison between our proposed RUL estimation models and already existing RUL estimation models.

\subsubsection{Performance Metrics for Evaluating RUL Estimation Models\label{sec:metrics}}
There are several metrics proposed to evaluate the performance of prognostics models (\cite{saxena2008metrics}). We measure the performance of our models in terms of Timeliness Score (S) and Root Mean Squared Error (RMSE):
For a test instance $i$, the error in estimation is given by $e_{i} =\hat{r}_{i} - {r}_{i}$. 
The timeliness score for $N$ test instances is given by
 $S=\sum^N_{i=1} (exp({\gamma\cdot|e_{i}|})-1)$, 
where $\gamma=1/\tau_1$ if $e_{i}<0$, else $\gamma=1/\tau_2$. 
Usually, $\tau_1 > \tau_2$ such that late predictions are penalized more compared to early predictions. 
We use $\tau_1=13$ and $\tau_2=10$ as proposed in \cite{saxena2008damage}.
The lower the value of $S$, the better is the performance.
The root mean squared error (RMSE) is given by:
$RMSE=\sqrt{\frac{1}{N}\sum^N_{i=1}e_{i}^2}$.

\begin{table}[h]
\centering
\footnotesize
\caption{Number of train, validation and test instances. Here, OC: number of operating conditions, FM: number of fault modes. \label{tab:datastats}}
\begin{tabular}{|c|c|c|c|c|c|c}
 \hline
{\bfseries Dataset}&{\bfseries Train}&{\bfseries Validation }&{\bfseries Test}&{\bfseries OC}&{\bfseries FM}\\
\hline
FD001&80&20&100&1&1\\
\hline
FD004&199&50&248&6&2\\
\hline
\end{tabular}
\end{table}

\begin{table}[h]
\centering
\footnotesize
\caption{Number of truncated instances. \label{tab:winstats}}
\begin{tabular}{|c|c|c|c|c}
 \hline
{\bfseries Dataset}&{\bfseries Train}&{\bfseries Validation }&{\bfseries Test}\\
\hline
FD001&1600&400&100\\
\hline
FD004&3980&1000&248\\
\hline
\end{tabular}
\end{table}

\begin{table*}[th]

\caption{Comparison of various LSTM-based approaches considered in terms of RMSE and Timeliness Score (S) for FD001 and FD004 datasets. 
$n_f$ and $n_c$ denote number of failed and censored instances in training set, respectively. \label{tab:ORvsMR}}
\scalebox{0.65}{%
\begin{tabular}{|c||c|c|c|c|c|c|c|c|c|c||c|c|c|c|c|c|c|c|c|c|}
\hline
\multicolumn{1}{|c||}{\bfseries} & \multicolumn{1}{|c}{} & \multicolumn{9}{c||}{\bfseries FD001}& \multicolumn{1}{|c}{} & \multicolumn{9}{c|}{\bfseries FD004}\\ \hline

 & \multicolumn{2}{|c|}{Instances}& \multicolumn{4}{|c|}{RMSE} & \multicolumn{4}{|c||}{Timeliness Score (S)} & \multicolumn{2}{|c|}{Instances} & \multicolumn{4}{|c|}{RMSE} & \multicolumn{4}{|c|}{Timeliness Score (S) $\times 10^{-3}$}\\ \hline

$\textit{p}_c(\%)$ & $n_f$ &$n_c$& MR & OR & ORC & ORCE & MR & OR & ORC & ORCE & $n_f$ &$n_c$ & MR & OR & ORC & ORCE &MR & OR & ORC& ORCE \\ \hline

0 & 80 & 0 & 15.62 & 15.63 & 15.63 & \textbf{14.62} & 507.2 & 367.64 & 367.64 & \textbf{292.76} & 199&0 &\textbf{26.88} & 28.33 & 28.33 & 27.47 & $\mathbf{4.92}$ & $6.44$ &$6.44$ & $5.24$\\ \hline

50 & 40 & 40 & 17.56 & 19.06& 17.60 & \textbf{15.98} & 444.1 &564.14 & 572.63 & \textbf{372.26} & 100&99 & \textbf{29.71} & 32.85& 31.48 & 30.62 & $7.97$ &$17.9$&$9.83$ & $\mathbf{7.86}$\\ \hline

70 & 24 & 56 & 19.92 & \textbf{16.48} & 18.53 & 16.57 & 713.31 & \textbf{362.21} & 561.11 & 404.94 & 60&139 & 33.17 & 33.65 &32.13 & \textbf{31.27} & $18.8$ &$17.4$ & $12.0$ & $\mathbf{9.59}$ \\ \hline

90 & 8 & 72 & 25.32 & 24.83 & 21.51 & \textbf{20.38} & $1.26\times 10^4$ & $3.07\times 10^4$ & $20.64\times 10^2$ & $\mathbf{13.57\times 10^2}$ & 20 & 179 & 41.23 & 43.88 & 39.75 & \textbf{38.41} & $102.0$ & $111.0$ & $141.13$ & $\mathbf{60.72}$ \\ \hline

\end{tabular}}
\end{table*}

\begin{figure*}[h]
\subfigure[RMSE FD001]{\includegraphics[trim={0cm 0cm 0cm 0cm},clip,width=0.24\textwidth]{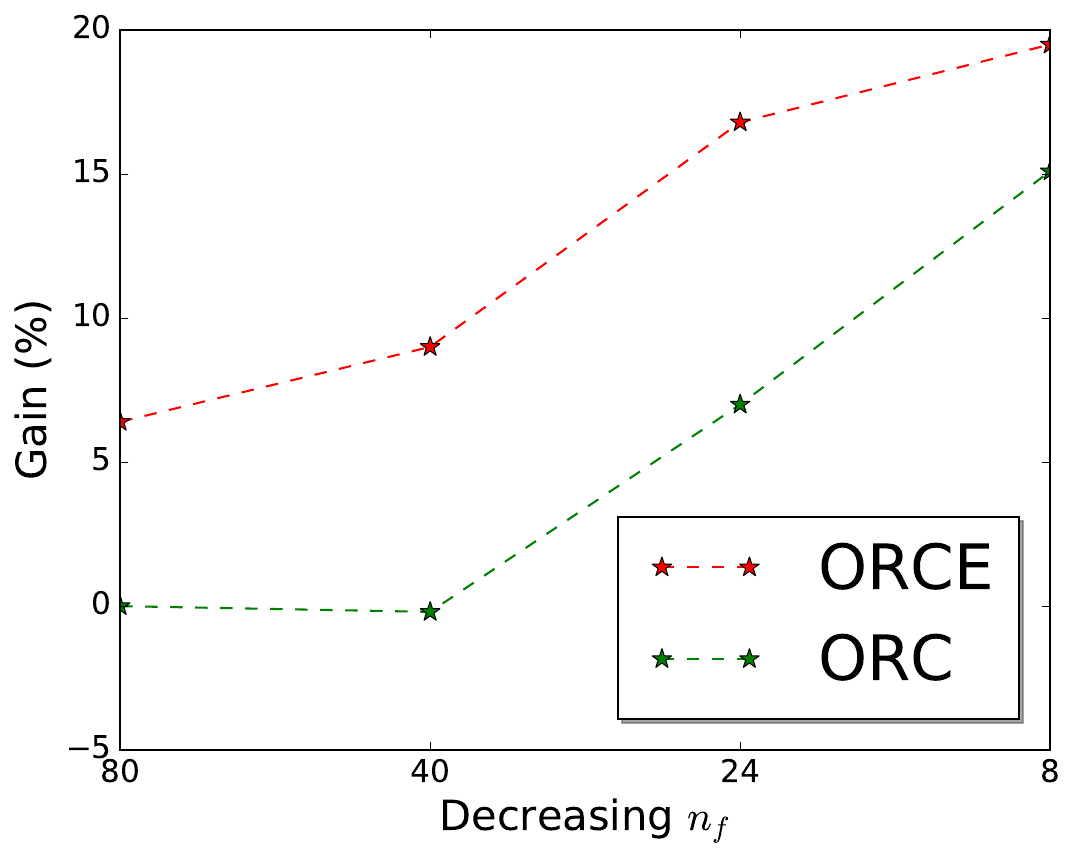}}
\subfigure[Timeliness Score (S) FD001]{\includegraphics[trim={0cm 0cm 0cm 0cm},clip,width=0.24\textwidth]{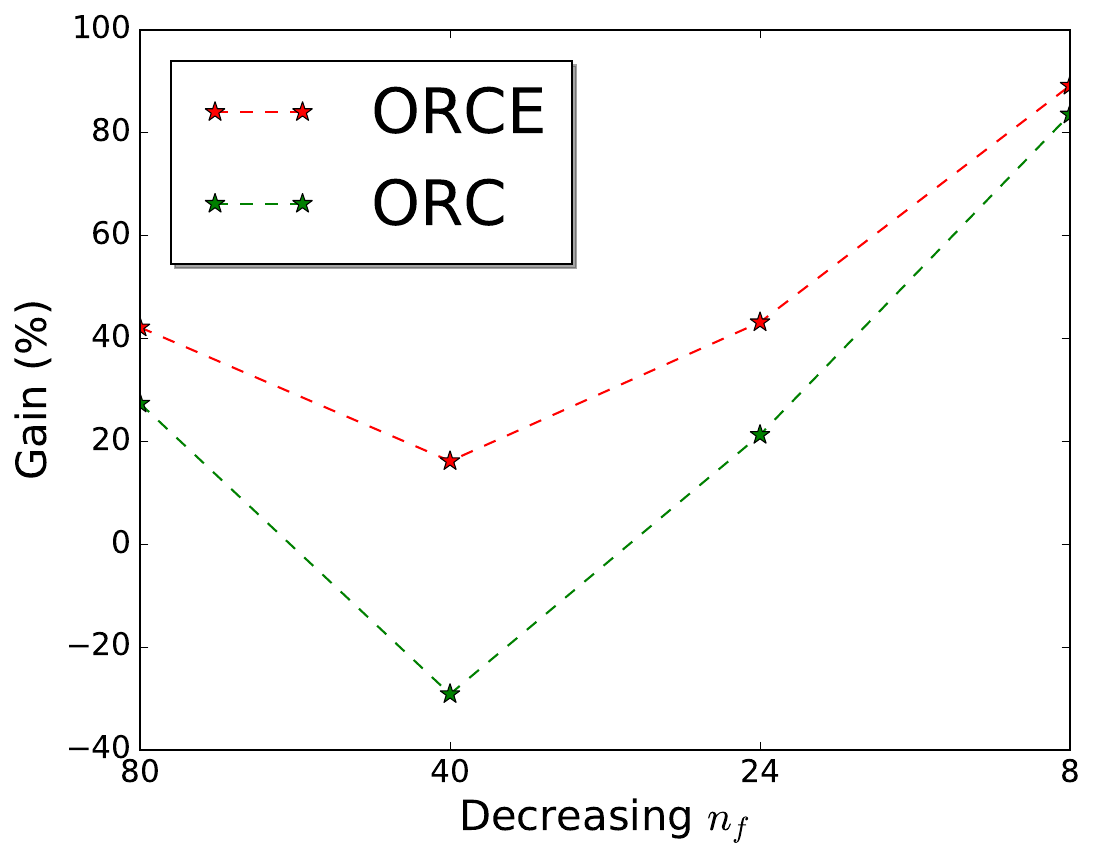}}
\subfigure[RMSE FD004]{\includegraphics[trim={0cm 0cm 0cm 0cm},clip,width=0.24\textwidth]{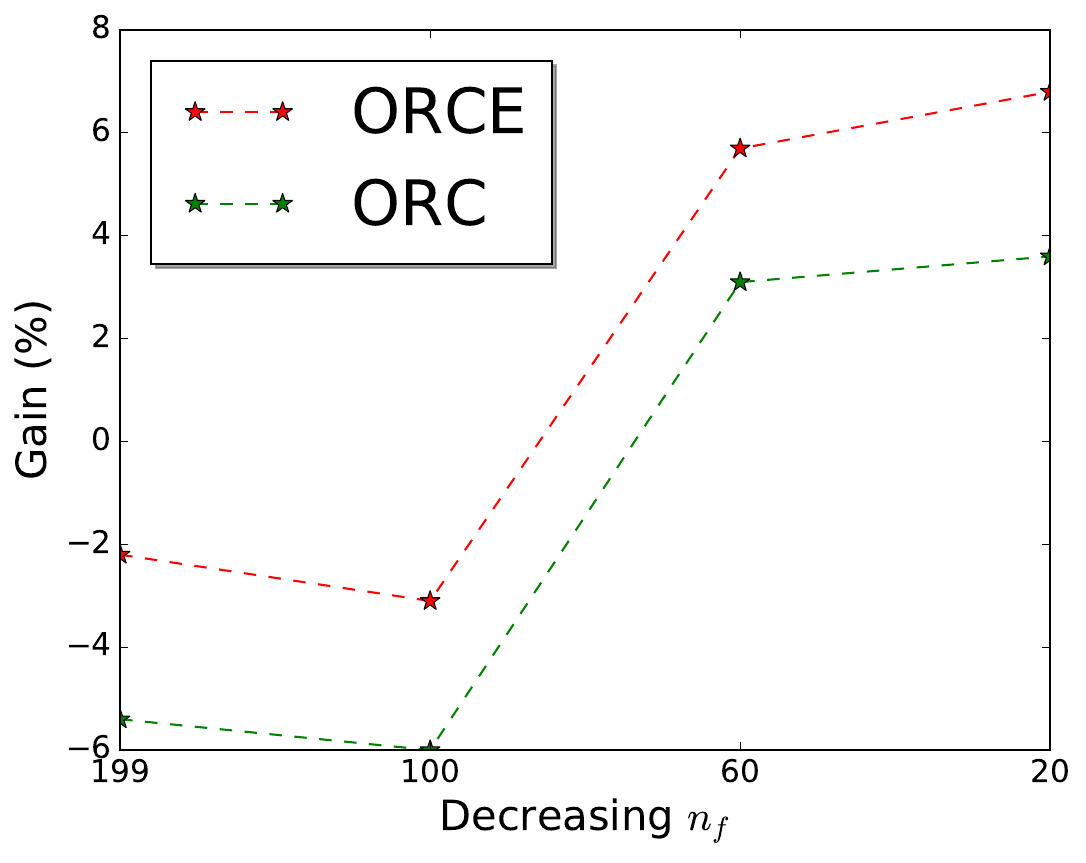}}
\subfigure[Timeliness Score (S) FD004\label{fig:ScoreFD004}]{\includegraphics[trim={0cm 0cm 0cm 0cm},clip,width=0.24\textwidth]{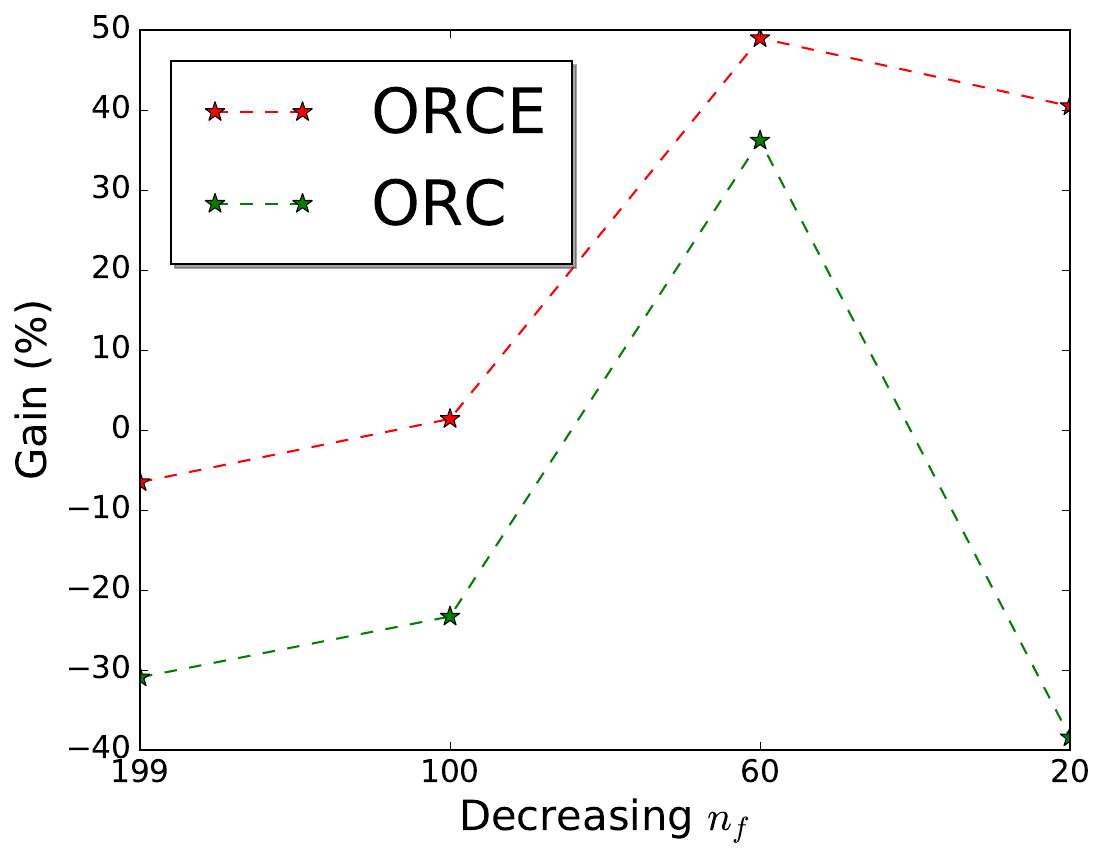}}

\caption{\%age gain of ORC and ORCE over MR with decreasing number of failed instances ($n_f$) in training.\label{fig:gain}}
\end{figure*}

\begin{figure*}[h]
	\subfigure[FD001\label{fig:PnRFD001}]{\includegraphics[trim={0.75cm 0cm 1.5cm 0cm},clip,width=0.24\textwidth]{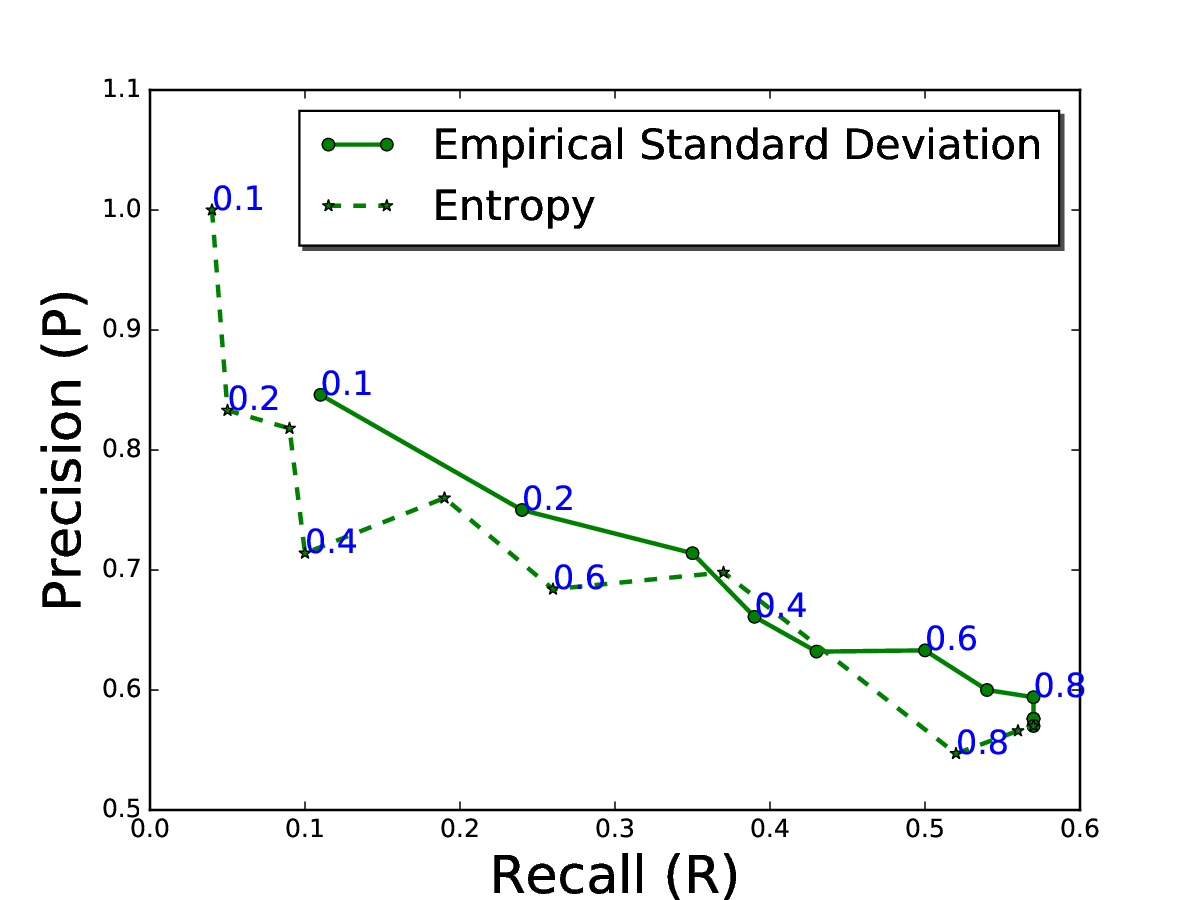}}
	\subfigure[FD004\label{fig:PnRFD004}]{\includegraphics[trim={0.75cm 0cm 1.5cm 0cm},clip,width=0.24\textwidth]{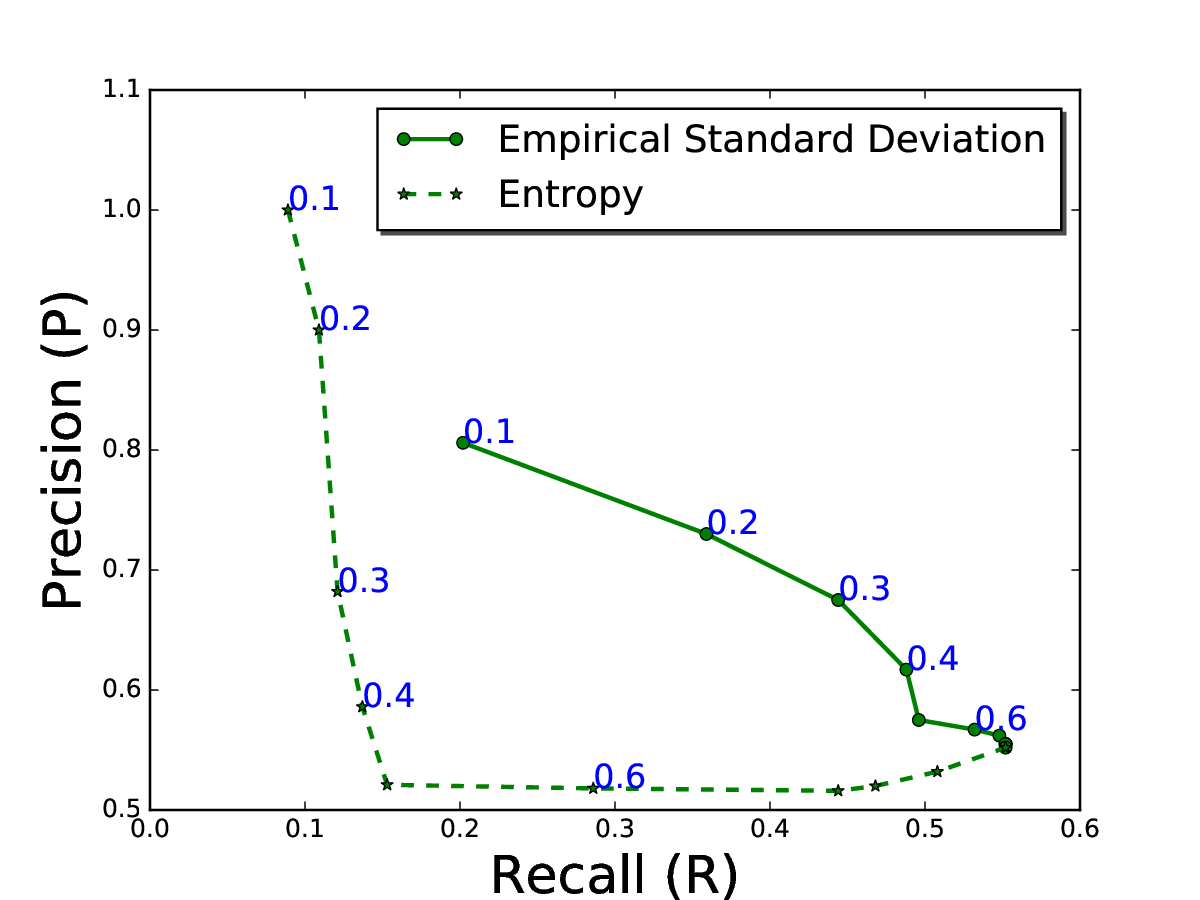}}
	\subfigure[FD001\label{fig:F1FD001}]{\includegraphics[trim={0.75cm 0cm 1.5cm 0cm},clip,width=0.24\textwidth]{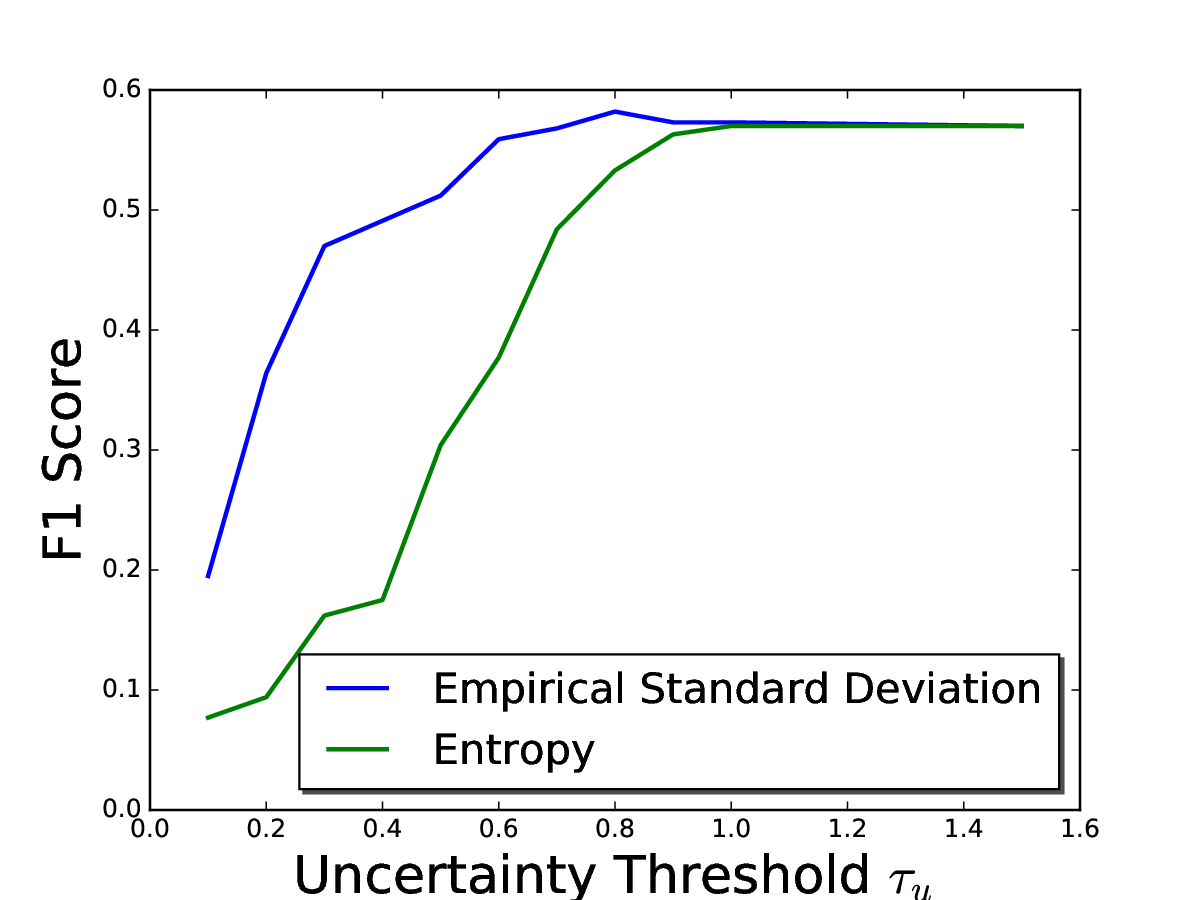}}
	\subfigure[FD004\label{fig:F1FD004}]{\includegraphics[trim={0.75cm 0cm 1.5cm 0cm},clip,width=0.24\textwidth]{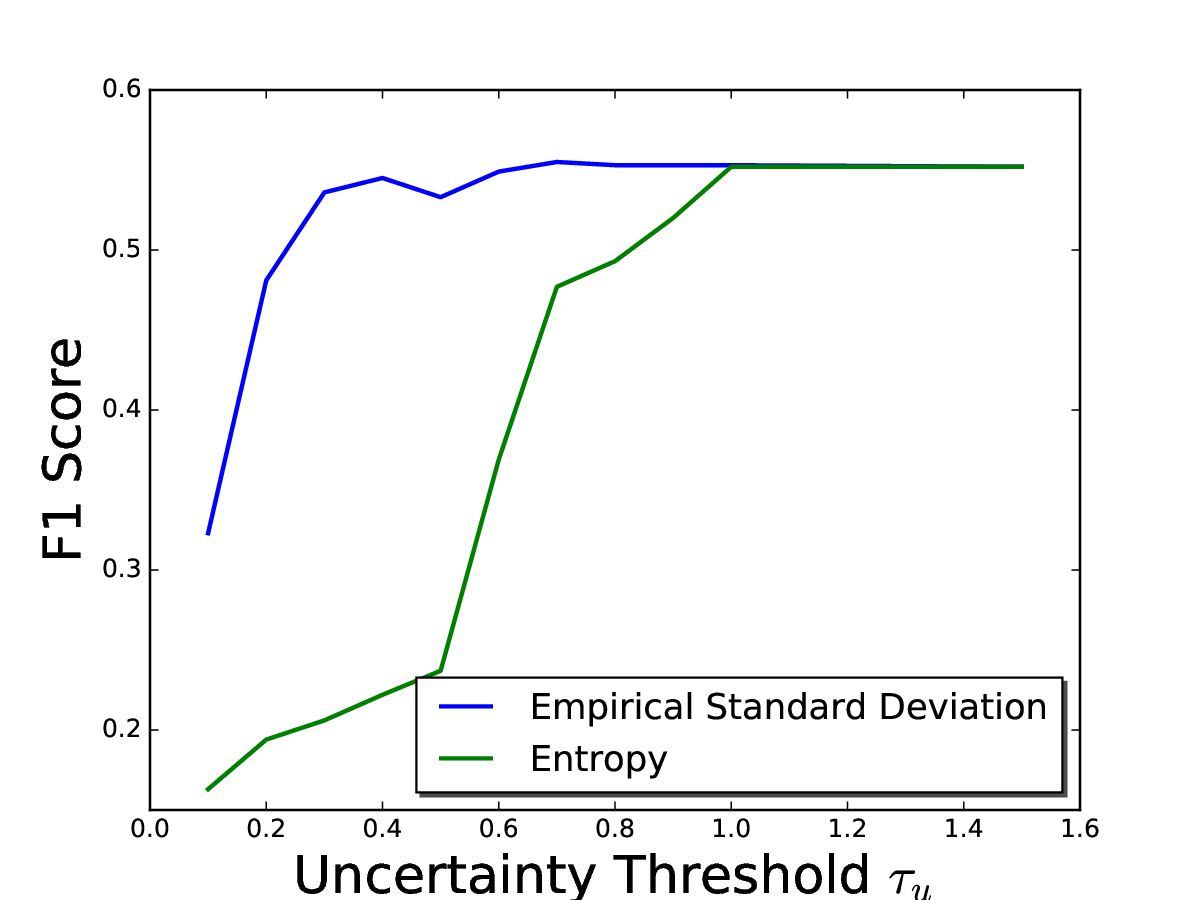}}
	\caption{Comparison of ESD and ENT as measures of uncertainty in terms of (a)-(b) Precision Recall Curves; and (c)-(d) F1 Scores with varying $\tau_u$. ESD is a more robust uncertainty metric compared to ENT.}
\end{figure*}

\begin{figure*}[h]
	\subfigure[Average Error with varying uncertainty threshold.\label{fig:sigma-error-plots-fd001-fd004}]{\includegraphics[trim={0cm 0cm 0cm 0cm},clip,width=\columnwidth]{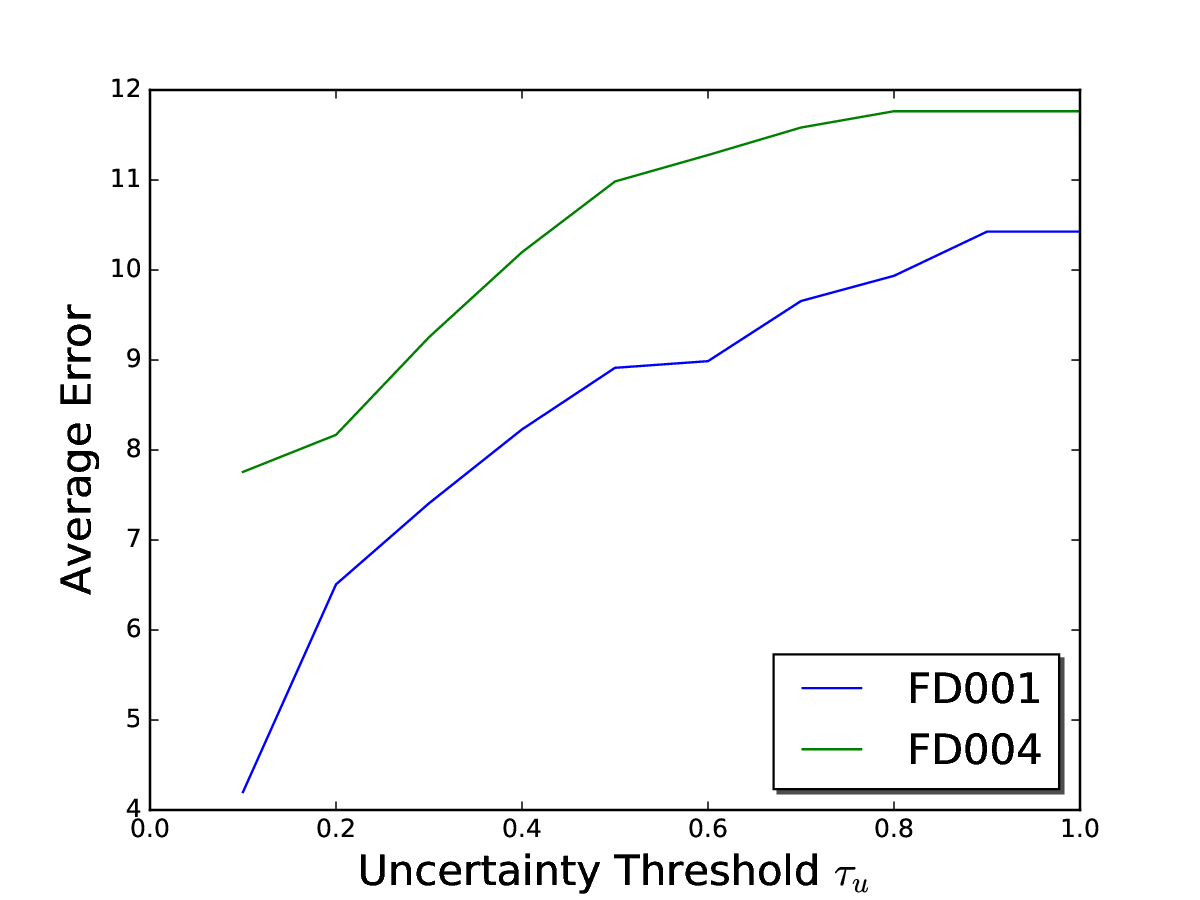}}
	\subfigure[Uncertainty Evaluation with varying RUL.\label{fig:ground-sigma-error-precision}]{\includegraphics[trim={0cm 0cm 0cm 0cm},clip,width=\columnwidth]{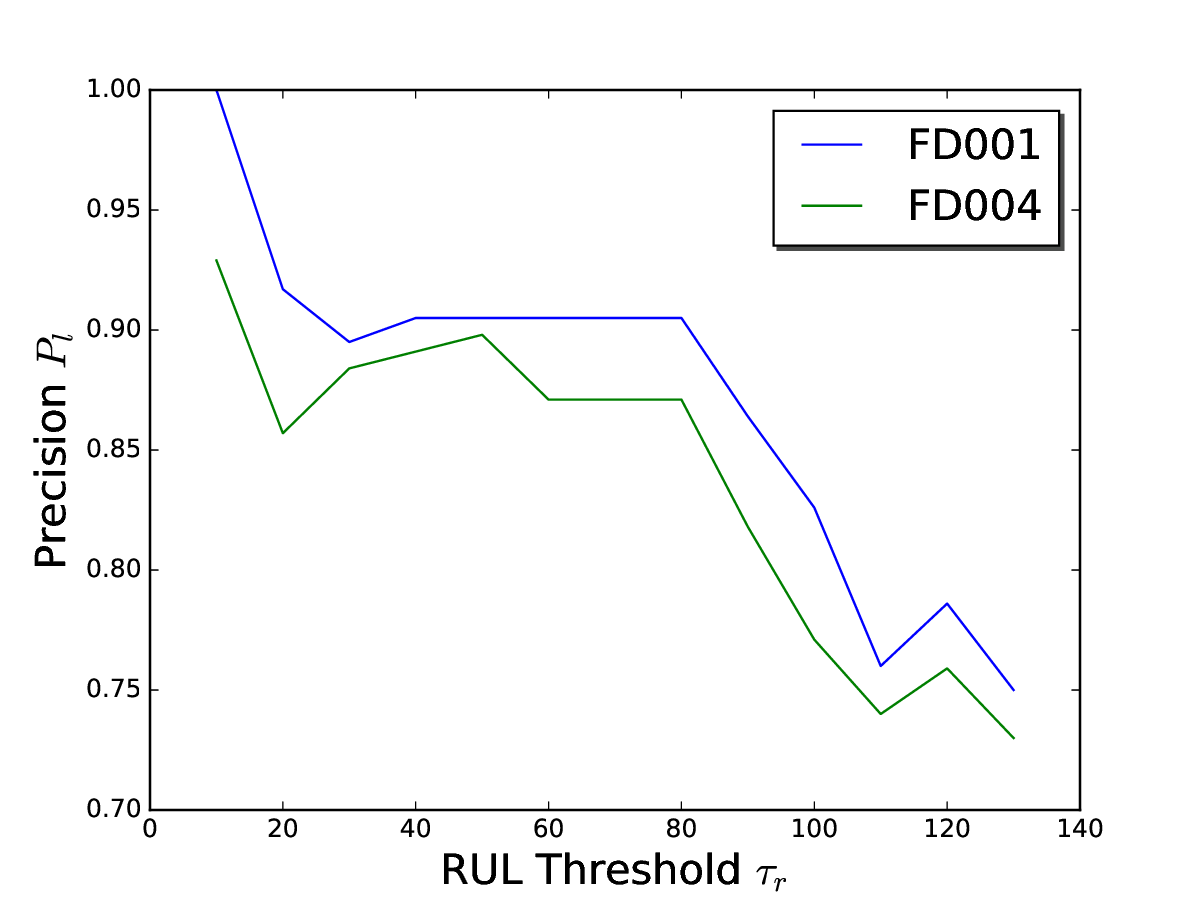}}
	\caption{Performance evaluation of ESD as an uncertainty metric showing: (a) lower uncertainty values corresponding to low RUL estimation errors, (b) highly precise and correct uncertainty estimates close to failures, i.e. when RUL is low.}
\end{figure*}

\subsubsection{Experimental Setup}

We consider $r_u=130$ cycles for all models, as used in \cite{babu2016deep,zheng2017long}.
For OR and ORC, we consider $K=10$ such that interval length $c=13$. 
For training the MR models, a normalized RUL in the range 0 to 1 (where 1 corresponds to a target RUL of 130 or more) is given as the target for each input. 
We use a maximum time series length of $T=360$; for any instance with more than 360 cycles, we take the most recent 360 cycles.
Also, we use the standard z-normalization to normalize the input time series sensor wise using mean and standard deviation of each sensor from the train set.

The hyperparameters $h$ (number of hidden units per layer), $L$ (number of hidden layers) and the learning rate are chosen from the sets $\{50,60,70,80,90,100\}$, $\{2,3\}$ and $\{0.001, \\
0.005\}$ respectively. We use a dropout rate of 0.2 for regularization, and a batch size of 32 during training. The models are trained for a maximum of 2000 iterations with early stopping.
The best hyperparameters are obtained using grid search by minimizing the respective loss function on the validation set.

For ORCE, we consider an ensemble of $m=6$ models (we consider up to 10 models in the ensemble, and found $m=6$ to work best across the scenarios considered).
The models are trained on the best hyperparameters selected from the corresponding hyperparameter sets of ORC.
While training different models, we ensure random initializations of the parameters of neural network and random shuffling of the training instances.
For selecting $m=6$ models from available $10$ models, we ordered the models in the ascending order of their respective loss values on the validation set and then select the first 6 models.

\subsubsection{Results and Observations\label{sec:ro_RUL}}

As summarized in Table \ref{tab:ORvsMR}, we observe that:
As the number of failed training instances ($n_f$) decreases, the performance for all models degrades (as expected).
However, importantly, for scenarios with small $n_f$, ORCE significantly outperforms MR and OR.   
For example, as shown in Figure \ref{fig:gain}, with $p_c=90\%$ (i.e. with $n_f=8$ and 20 for FD001 and FD004, respectively), ORCE performs significantly better than MR, and shows 19.5\% and 6.8\%  improvement over MR in terms of RMSE, for FD001 and FD004, respectively.
The gains in terms of timeliness score $S$ are higher because of the exponential nature of $S$ (refer Section \ref{sec:metrics}).
It is evident from Figure \ref{fig:gain} that ORCE is performing better than ORC and MR in terms of both RMSE and S. 
The performance gap between ORCE and ORC significantly increases in case of timeliness score (S) for FD004 dataset when $p_c=90\%$, shown in \ref{fig:ScoreFD004}.
Due to fewer number of failed training instances ($p_c=90\%$), some models in the ensemble are not trained properly and result in high errors even for the instances with lower RUL $r$. This results in very high values of $S$. 
In case of ORC, the overall value of $S$ tends to be high since, for ORC, we compute the average of timeliness scores corresponding to $m$ models in an ensemble. This is not the case for ORCE, since the instance-wise RUL estimations are obtained as the average of $m$  estimations from the $m$ models in the ensemble, the performance of ORCE in terms of $S$ is better when compared to ORC. 

While MR and OR have access to only a small number failed instances $n_f$ for training, ORCE and ORC have access to $n_f$ failed instances as well as partial labels from $n_c$ censored instances for training. 
Therefore, MR and OR models tend to overfit while ORC and ORCE models are more robust.
 
We also provide a comparison with existing deep CNN-based (\cite{babu2016deep}) and LSTM-based (\cite{zheng2017long}) MR approaches in Table \ref{tab:litComp}. 
ORC (same as OR for $m=$0\%) performs comparably to existing MR methods.
More importantly, as noted above, ORC and ORCE may be advantageous and more suitable for practical scenarios with few failed training instances.

\subsection{Uncertainty Quantification\label{sec:uncertainty}}
We introduce various metrics used to evaluate the performance of the proposed ensemble-based uncertainty estimation approach.
Using these metrics, we demonstrate the efficacy of the proposed approach from a practical point of view.
We compare the proposed ESD (Equation \ref{eq:uncertaintyEstStdDev}) and two variants of entropy (as introduced in Appendix \ref{apx:ent_uncertainty}) for uncertainty evaluation. 

\subsubsection{Performance Metrics for Evaluating Uncertainty \\Quantification Methods\label{sec:precision}}
We expect our model to be certain (have high certainty) when the RUL estimates are correct, and less certain (have low certainty) for highly erroneous RUL estimates. 
We consider an RUL estimation to be correct if the absolute error $|r - \hat{r}| \leq \tau_e$, and to be certain if the corresponding uncertainty estimate $\hat{u} \leq \tau_u$.
Also, for evaluating the performance of uncertainty metrics we restrict the target RUL $r$ to a maximum of $r_u=130$ because we train our models with a maximum target RUL of $r_u$ and so $\hat{r}$ cannot be greater than $r_u$. This is done because even if the model confidently estimates $\hat{r}$ close to $r_u$, a value of $r$ much greater than $r_u$ will lead to high error and cannot result in proper performace evaluation of the uncertainty metrics.
Under above considerations, we measure precision and recall to evaluate the performance of uncertainty quantification approach as follows: 
Precision is the fraction of test instances with uncertainty below a threshold $\tau_u$ that also have error $\leq \tau_e$. Recall is defined as the fraction of test instances having uncertainty and error below some threshold $\tau_u$ and $\tau_e$, respectively. More specifically:

\begin{equation}\label{eq:uncertaintyEstPrecison}
\begin{aligned}
Precision(P)&= \frac{\#(\hat{u} \leq \tau_u) \cap \#(|r - \hat{r}| \leq \tau_e)}{\#(\hat{u} \leq \tau_u)},\\
Recall(R)&= \frac{\#(\hat{u} \leq \tau_u) \cap \#(|r - \hat{r}| \leq \tau_e)}{\#(test\ instances)}, \\
F1&= 2\times \frac{P \times R} {P+R}
\end{aligned}
\end{equation}
where $\#(X)$ denotes the number of instances satisfying the condition $X$.

Further, it is desirable to have very certain and correct estimates close to failure to avoid fatal consequences upon failure.
To evaluate performance from this point-of-view, we analyze the relation of uncertainty with nearness to failure.
It is desirable to have low error as well as low uncertainty when $r$ is low. To evaluate this aspect, we study the variation in precision for different RUL thresholds $\tau_r$, considering test instances with low ground truth RULs. The modified precision $P_{l}$ in this context is given by:

\begin{equation}\label{eq:uncertaintyCoverage}  
\begin{aligned}
P_{l} = \frac{\#(r \leq \tau_r) \cap \#(\hat{u} \leq \tau_u) \cap \#(|r - \hat{r}| \leq \tau_e)}{\#(r \leq \tau_r) \cap \#(\hat{u} \leq \tau_u))}
\end{aligned}
\end{equation}
    
For given thresholds $ \tau_r$ and $\tau_u$, $P_l$ quantifies the fraction of test instances with actual RUL  $r \leq \tau_r$ and uncertainty $ \leq \tau_u$  that also have error $\leq \tau_e$ .

\begin{table}[th]
\caption{Performance comparison of the proposed approach with existing approaches in terms of RMSE and Timeliness Score (S).\label{tab:litComp}}
\scalebox{0.73}{%
\begin{tabular}{|c|c|c|c|c|}
\hline
\multicolumn{1}{|c|}{} & \multicolumn{2}{c}{\bfseries FD001} & \multicolumn{2}{|c|}{\bfseries FD004}\\ \hline
 & RMSE & S & RMSE & S\\ \hline

CNN-MR (\cite{babu2016deep}) & 18.45 &$1.29\times 10^3$ & 29.16 &$7.89\times 10^3$ \\ \hline

LSTM-MR (\cite{zheng2017long}) & 16.14 &$3.38\times 10^2$ &28.17 &$5.55\times 10^3$ \\ \hline

MR (ours) & 15.62 & $5.07\times 10^2$&\textbf{26.88} & $\mathbf{4.92\times 10^3}$\\ \hline

ORC (proposed) & 15.63 &$3.68\times 10^2$& 28.33 &$6.44\times 10^3$ \\ \hline
ORCE (proposed) & \textbf{14.62} & $\mathbf{2.93\times 10^2}$ & 27.47 & $5.24\times 10^3$ \\ \hline
\end{tabular}}
\end{table}

\subsubsection{Results and Observations}
For sake of brevity, we restrict the results and observations to the uncensored scenario, i.e. $p_c=0\%$. Similar results and observations for models corresponding to censored scenarios are presented in Appendix \ref{apx:detailed}.

\textit{Comparing ESD vs Entropy (ENT) as uncertainty metric}: Precision and Recall (as in Equation \ref{eq:uncertaintyEstPrecison}) are used to compare the two approaches for uncertainty estimation. Precision-Recall curves are obtained by varying the threshold on uncertainty $0.1\leq \tau_u \leq 1.5$ while keeping $\tau_e = 10$.
We observe that for $R \geq 0.1$, P is higher in case of ESD for FD001 dataset, shown in Figure \ref{fig:PnRFD001}.
Similar behavior is observed in case of FD004 dataset, for $R \geq 0.2$, shown in Figure \ref{fig:PnRFD004}.
We further plot $F1$ score (as in Equation \ref{eq:uncertaintyEstPrecison}) by varying the $\tau_u$, shown in Figure \ref{fig:F1FD001} and \ref{fig:F1FD004}, which shows that ESD is a better uncertainty quantification metric compared to ENT. (We also analyze the instances for which ESD has unexpected behavior in terms of low uncertainty while having high error in RUL estimate. The observations are given in Appendix \ref{apx:detailed}.)

\textit{Relation between uncertainty and error}: For a reliable model, RUL estimates with high certainty must be accurate, i.e. have low RUL estimation errors. 
To evaluate the performance of uncertainty metric in this context, we consider instances with uncertainty $\hat{u}\leq \tau_u$, and compute the average error in RUL estimation for these instances.
As shown in Figure \ref{fig:sigma-error-plots-fd001-fd004}, we observe that for low values of $\tau_u$, the average error thus computed is also low, indicating that the model is more accurate when it is more certain. 
Further, as expected, we observe an increase in average error with increasing $\tau_u$, suggesting that the RUL estimates tend to be more erroneous when the model is uncertain.
 
\textit{Relation between uncertainty and actual RUL}: 
For quantifying the relationship between RUL and uncertainty, $P_{l}$ is calculated as in Equation \ref{eq:uncertaintyCoverage}.
$P_{l}$ is computed for varying $\tau_r$, ranging from $10$  to $130$ and, keeping $\tau_u$ and $\tau_e$ fixed as $0.2$ and $10$ respectively.
From practical point of view, higher precision ($P_{l}$) in case of lower values of $\tau_r$ is expected to correctly and confidently handle instances that are approaching failure. Similar trend is observed in our case also, as shown in Figure \ref{fig:ground-sigma-error-precision}. For $\tau_r =20$, $P_l=0.917$ for FD001 dataset and $P_l=0.857$ for FD004 dataset suggests that the model is certain and accurate $91.7\%$ of the times for FD001 dataset and $85.7\%$ of the times for FD004 dataset.

\section{Conclusion and Discussion\label{sec:conc}}
In this work, we have proposed a novel approach for RUL estimation using deep ordinal regression based on multilayered LSTM neural networks.
We have argued that ordinal regression formulation is more robust compared to metric regression, as the former allows for incorporation of more labeled data from censored instances. 
We found that leveraging censored instances significantly improves performance when the number of failed instances is small.
In future, it would be interesting to see if a semi-supervised approach (e.g. as in \cite{yoon2017semi,gugulothu2018on}) with initial unsupervised pre-training of LSTMs using failed as well as censored instances can further improve the robustness of the models. Further, an extension to the proposed approach to address the usually encountered non-stationarity scenario using approaches similar to \cite{saurav2018online} can be considered.
It is to be noted that although we have experimented with LSTMs for Ordinal Regression, our OR approach is generic enough to be useful for any neural network, e.g. CNNs. 

Further, we have proposed a simple yet effective approach to quantify uncertainty in the RUL estimates by using a simple average ensemble of the deep ordinal regression models. The proposed empirical standard deviation based metric for uncertainty provides accurate predictive uncertainty estimates: we observe low errors in RUL estimation for low uncertainty values. Further, the model is found to be accurate with high certainty when the remaining useful life is very low, i.e. the instance is approaching failure. It will be interesting to see if the ensemble based approach for uncertainty quantification can be extended to metric regression models as well using uncertainty methods for regression as proposed in \cite{NIPS2017_7219}.

\bibliographystyle{apacite}
\PHMbibliography{BibTex/phm-kdd2016,BibTex/phm-kdd2017,BibTex/sensor_analytics,BibTex/online-ad,BibTex/ijcai2017,BibTex/kdd2018,BibTex/ecml-pkdd2018,BibTex/dise}

\appendix
\section{Appendix\label{apx:appendix}}

\subsection{Entropy as a measure of uncertainty\label{apx:ent_uncertainty}}
Let $\mathcal{Z}$ set of possible values of the multi-label target vector. In our case, since we are using OR, the number of possible combinations for any given $K$ is $K+1$. For example, when $K=3$, $\mathcal{Z} = \{000,001,011,111\}$.

For entropy, we average out the $m$ estimates to get the final $\hat{\mathbf{y}} = \sum_{i=1}^m\hat{\mathbf{y}}_i$. The uncertainty for a test instance, considering the independent nature of all of the $K$ binary classifier is given as follows:
\begin{equation}\label{eq:uncertaintyEstEntropy}
\begin{aligned}
\hat{u}_{ENT} = -\sum_{\mathbf{z}\in \mathcal{Z}}P(\mathbf{z})\ log(P(\mathbf{z})) \\
P(\mathbf{z}) = \prod_{j=1}^{K} P(z_{j}) \\
P(z_{j}) = 
\begin{cases}
\hat{y}_{j} & z_j = 1\\
1 - \hat{y}_{j} & z_j = 0\\
\end{cases}
\end{aligned}
\end{equation}

We normalize the uncertainty values ($\hat{u}_{ENT}$) using the minimum and maximum uncertainty values across all instances in a hold-out validation set through min-max normalization.
For entropy calculations, we also tried by calculating the entropy for each of the $K$ classifiers individually and later averaging them out to get the final one. But, it was less effective as compared to the approach defined before.

\begin{figure*}[h]
\subfigure[FD001 Dataset]{\includegraphics[trim={0cm 0cm 0cm 0cm},clip,width=0.4\textwidth]{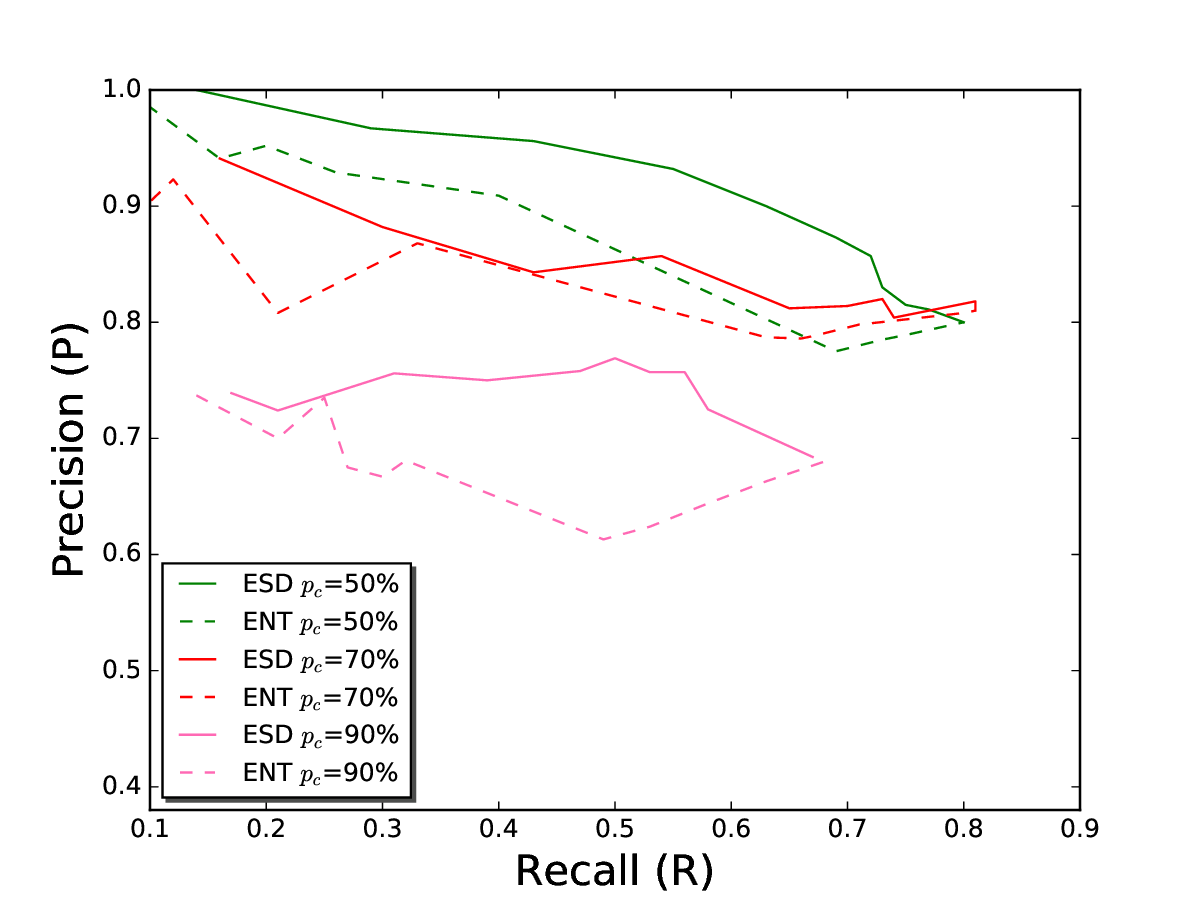}}\hspace{0.1\textwidth}
\subfigure[FD004 Dataset]{\includegraphics[trim={0cm 0cm 0cm 0cm},clip,width=0.4\textwidth]{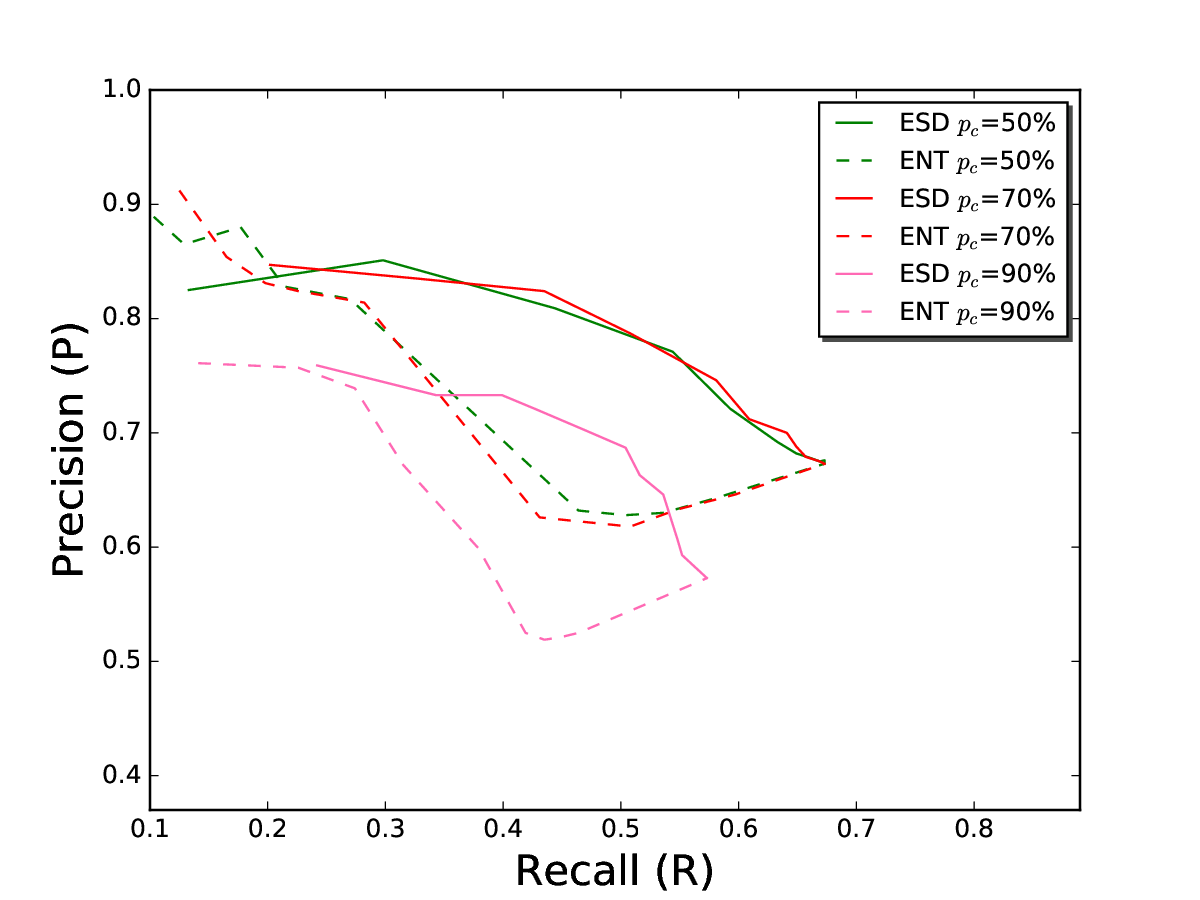}}
\caption{Precision Recall Curves comparing ESD and ENT.\label{fig:entropy-deviation-precision-recall}}
\end{figure*}

\begin{figure*}[h]
\subfigure[FD001 Dataset]{\includegraphics[trim={0cm 0cm 0cm 0cm},clip,width=0.4\textwidth]{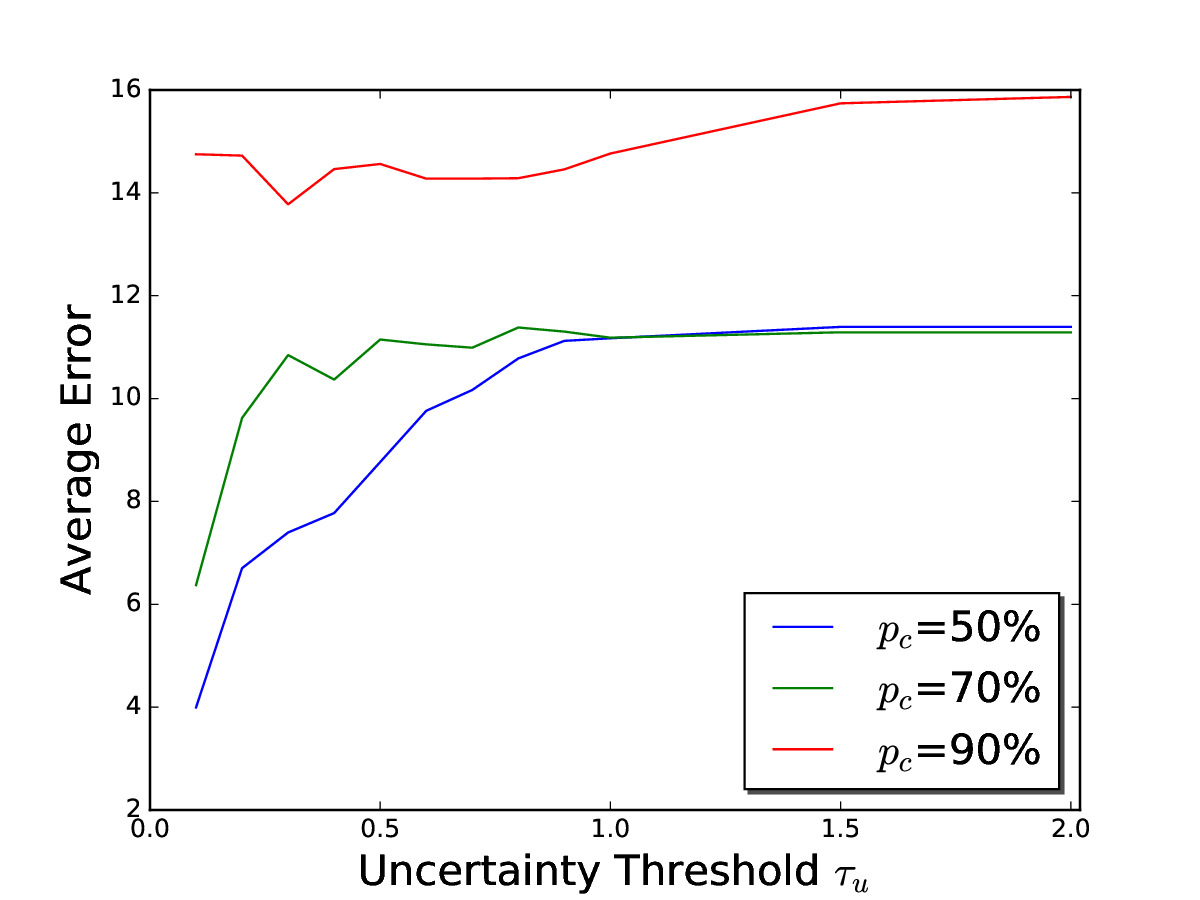}}\hspace{0.1\textwidth}
\subfigure[FD004 Dataset]{\includegraphics[trim={0cm 0cm 0cm 0cm},clip,width=0.4\textwidth]{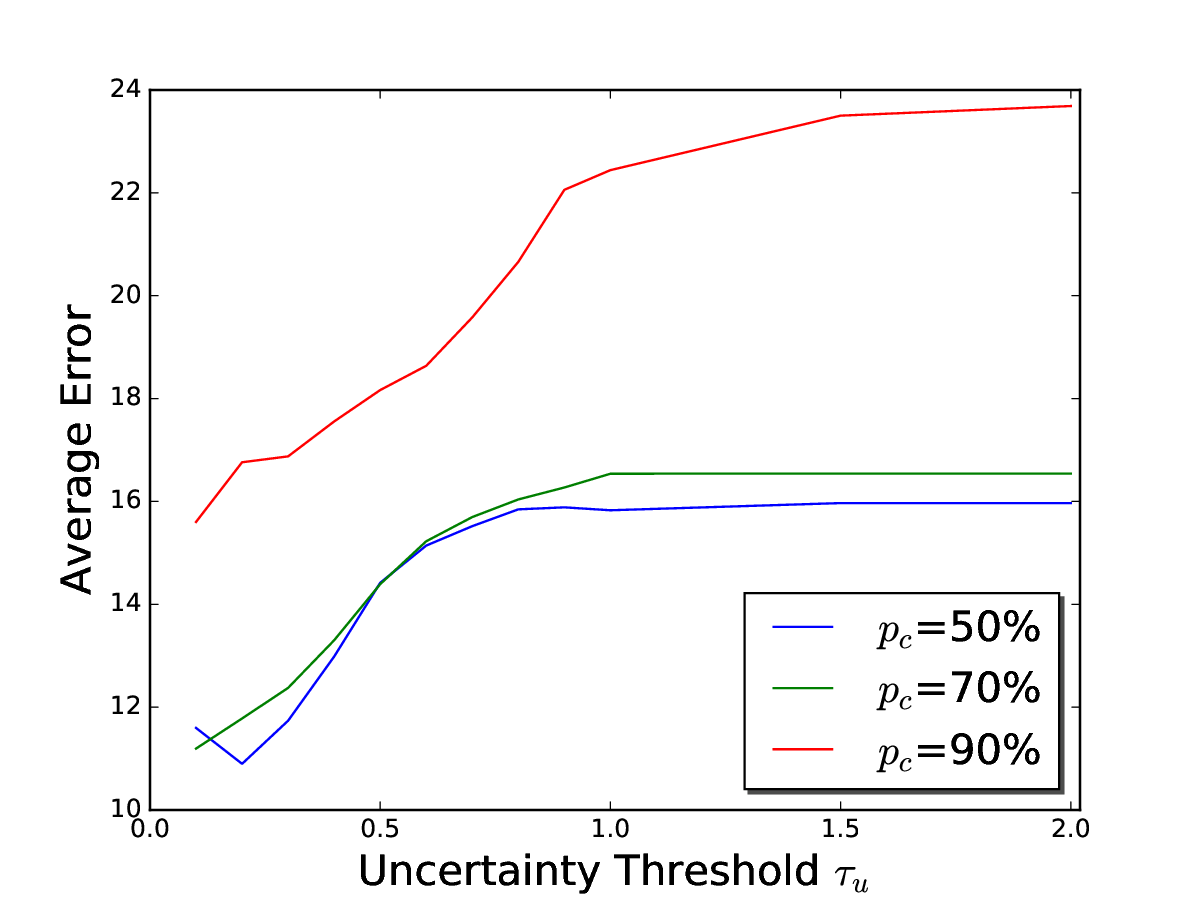}}
\caption{Average Error at varying $\tau_u$.\label{fig:sigma-error-plots-new}}
\end{figure*}

\begin{figure*}[h]
\subfigure[FD001 Dataset]{\includegraphics[trim={0cm 0cm 0cm 0cm},clip,width=0.4\textwidth]{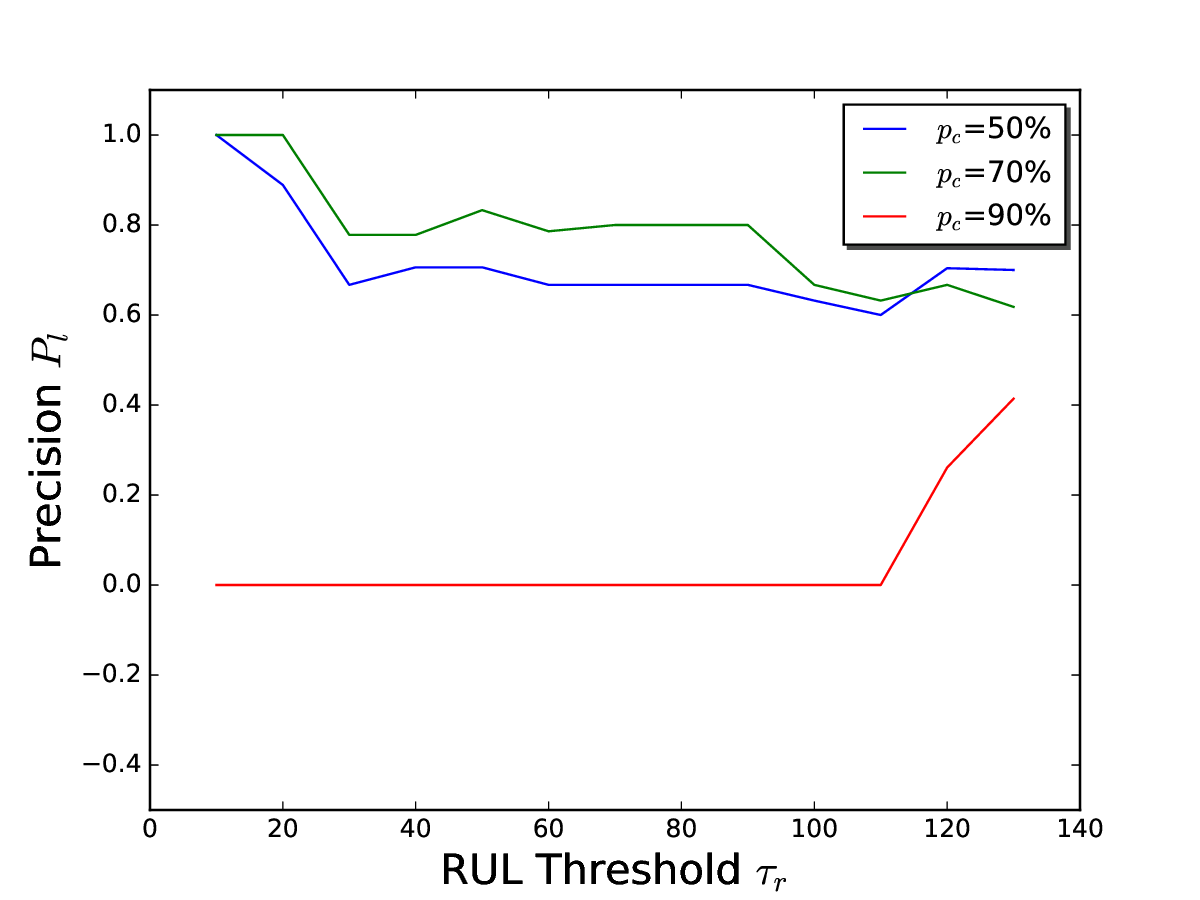}}\hspace{0.1\textwidth}
\subfigure[FD004 Dataset]{\includegraphics[trim={0cm 0cm 0cm 0cm},clip,width=0.4\textwidth]{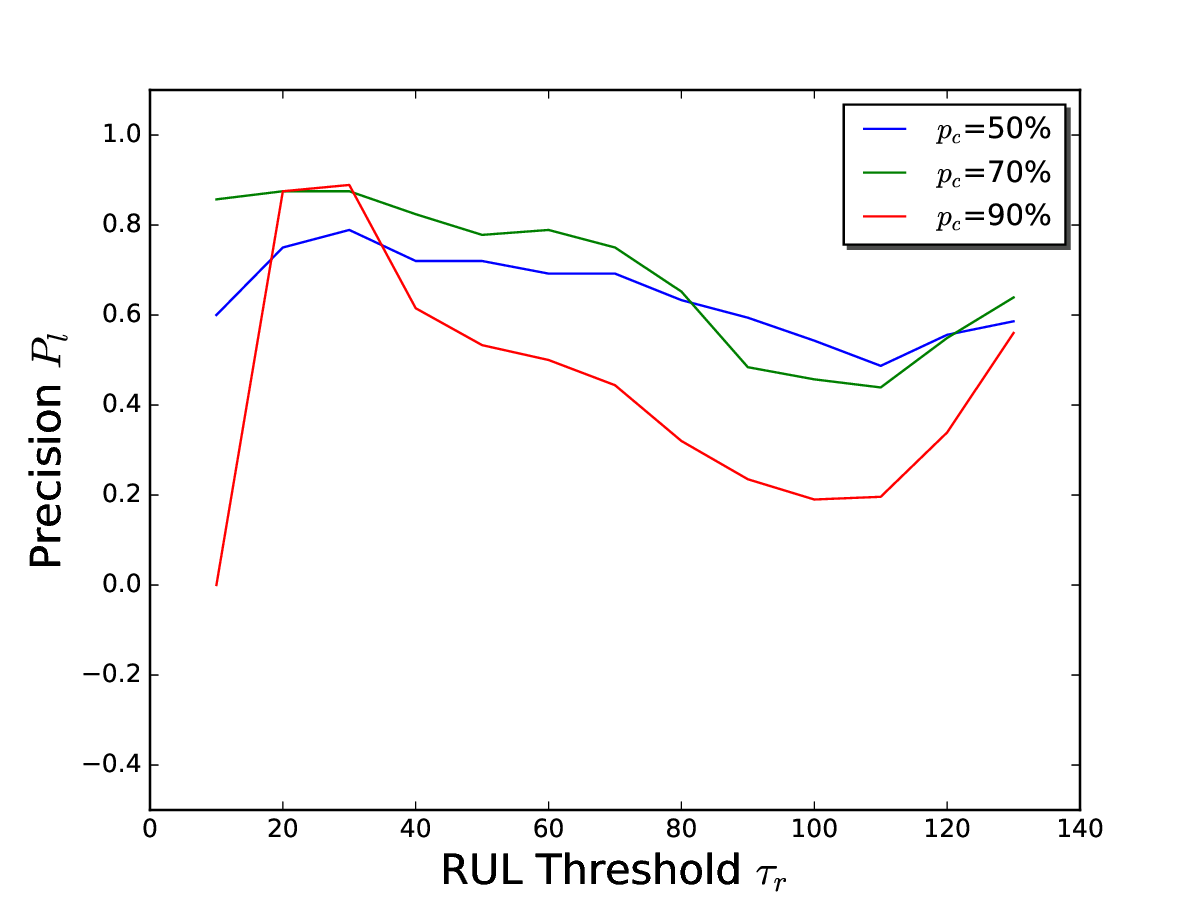}}
\caption{Uncertainty Evaluation wrt RUL.\label{fig:ground-coverage}}
\end{figure*}

\subsection{Detailed Evaluation for Uncertainty Quantification\label{apx:detailed}}

\textit{Instance Level Analysis} : We perform a qualitative analysis of test instances having low uncertainty values despite having high error values to understand the scenarios where our proposed uncertainty quantification measure is failing. 
For such test engine (test instance), we consider three nearest engines from the training set where nearness is defined in terms of Euclidean distance between the target vector estimate $\hat{\mathbf{y}}$ of test and training engines. 
After finding the nearest training engines, we plot the first PCA component corresponding to the test and respective nearest training engines. 

For PCA process, each multivariate reading from each timestamp is reduced to univariate reading by taking the first principal component as discussed in \cite{malhotra2016multi}.

\begin{figure*}[h]
\subfigure[FD001 Dataset]{\includegraphics[trim={0cm 0cm 0cm 0cm},clip,width=0.4\textwidth]{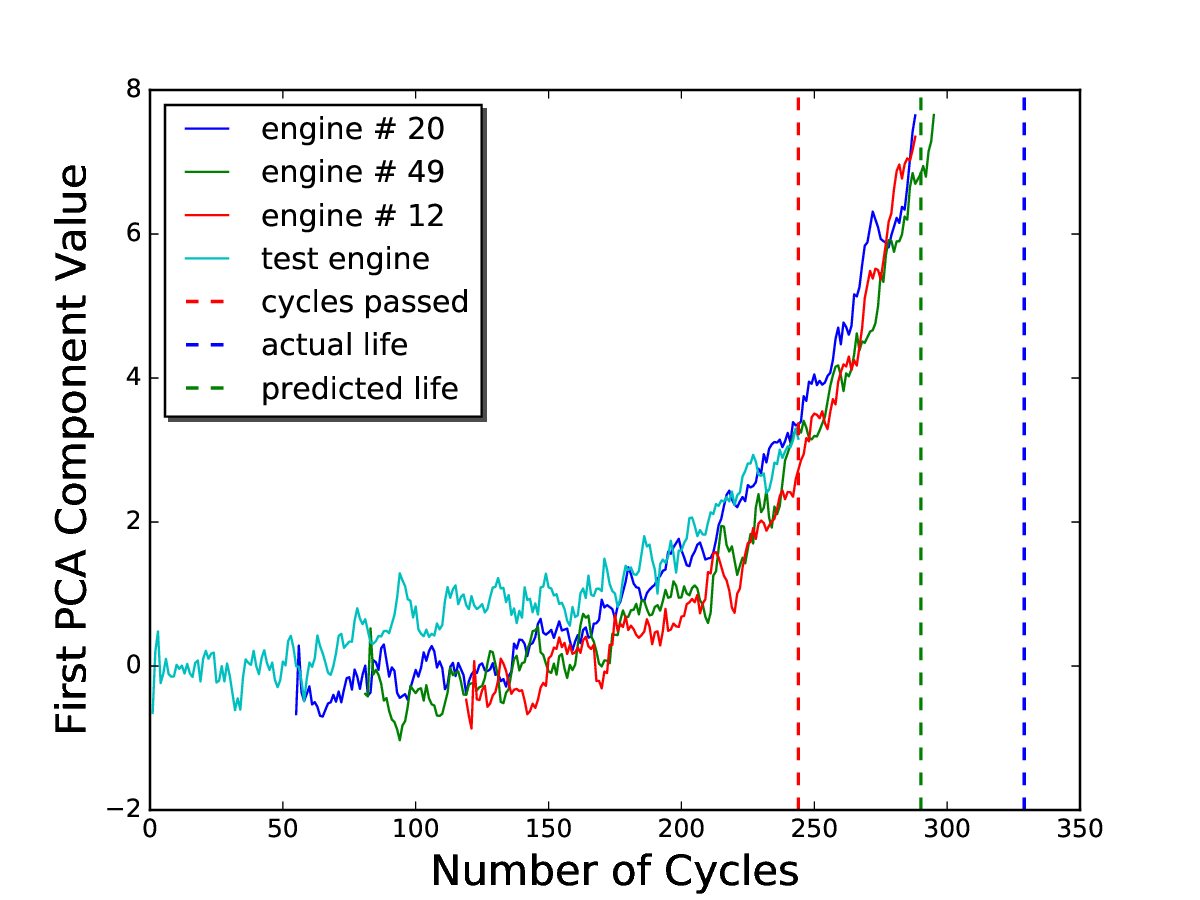}}\hspace{0.1\textwidth}
\subfigure[FD004 Dataset]{\includegraphics[trim={0cm 0cm 0cm 0cm},clip,width=0.4\textwidth]{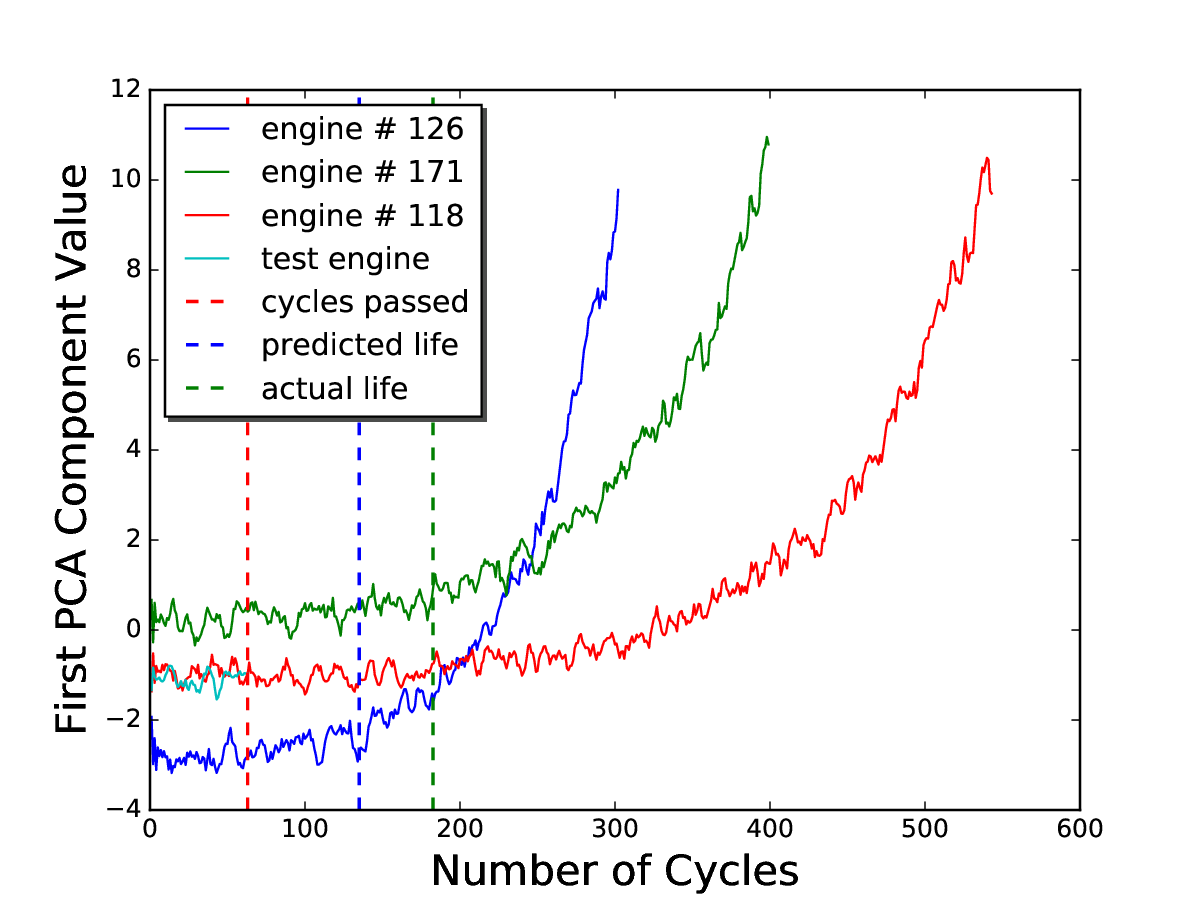}}
\caption{Instance Level Analysis (PCA) showing instances with low uncertainty estimate but high error in RUL estimate.\label{fig:pcafig}}
\end{figure*}

PCA plot for test engine\#93 from FD001 is shown in Figure \ref{fig:pcafig}(a).
Although a large number of cycles have passed for this instance, it has a significantly high RUL. The total life of this instance is $329$ making it a rare instance as instances with such high total cycles are not observed in the training data.
Despite the passage of higher number of cycles, RUL is very high. Training engines with such high RUL are rare which is leading to the higher error inspite of having low uncertainty. 

Similarly, PCA plot for test engine\#166 from FD004 is shown in Figure \ref{fig:pcafig}(b).
Due to the assumption of having maximum RUL $r$ as $130$ in ORC formulation, predicted life is around $130$, even though the actual life is significantly higher.
This clipping effect causes the high error. Moreover, scarcity of training engines with such high RUL further leads to increase in error inspite of having low uncertainty.

\textit{Uncertainty Evaluation wrt Error}: We expect our model to be highly uncertain for higher error values. To evaluate the same, we calculate $C_{e}$ as follows:
\begin{equation}\label{eq:uncertaintyEstErrorCoverage}
C_{e} = \frac{\#(|r - \hat{r}| \leq \tau_e) \cap \#(\hat{u} \leq \tau_u)}{\#(|r - \hat{r}| \leq \tau_e)}
\end{equation}
where, $C_e$ is the fraction of test instances with error and uncertainty $\leq \tau_e$ and $\leq \tau_u$ respectively.
\begin{figure*}[h]
\subfigure[FD001 Dataset]{\includegraphics[trim={0cm 0cm 0cm 0cm},clip,width=0.4\textwidth]{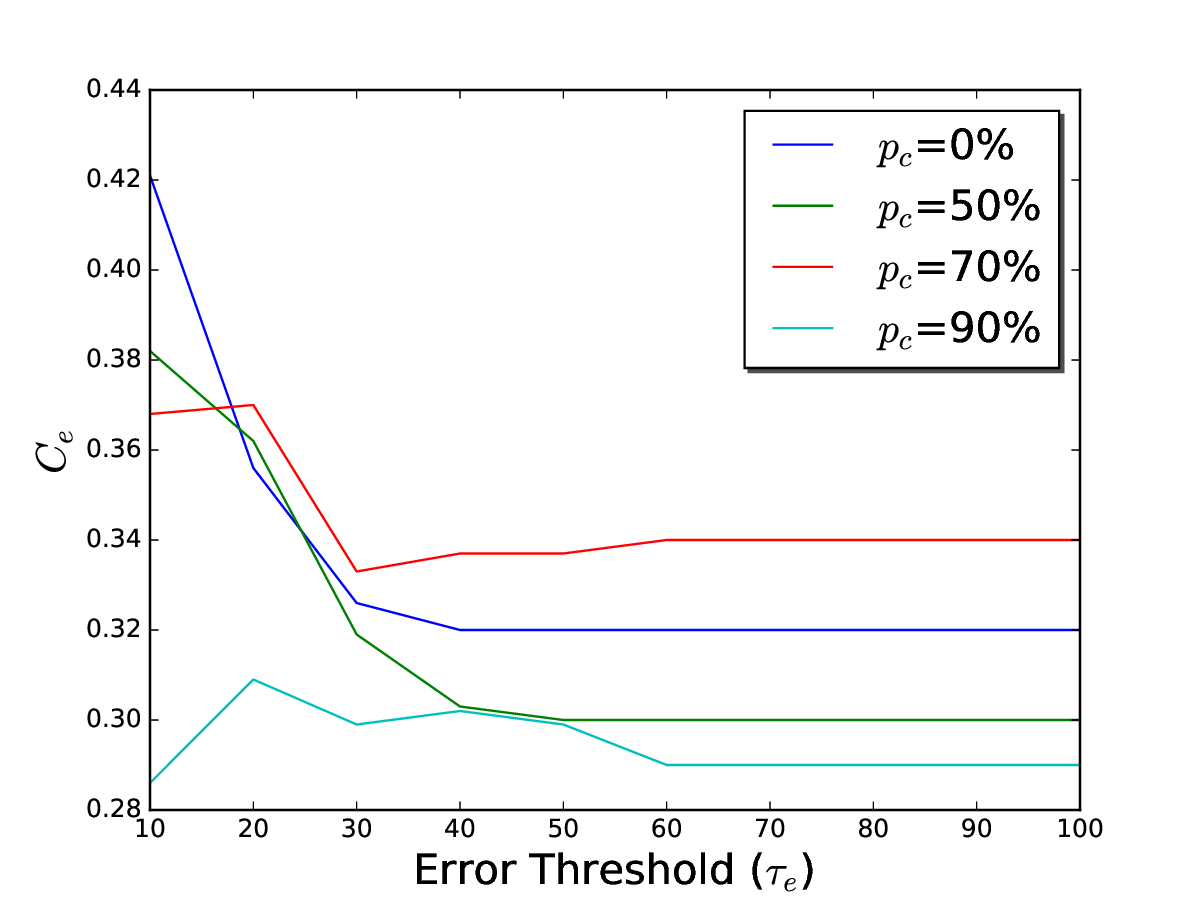}}\hspace{0.1\textwidth}
\subfigure[FD004 Dataset]{\includegraphics[trim={0cm 0cm 0cm 0cm},clip,width=0.4\textwidth]{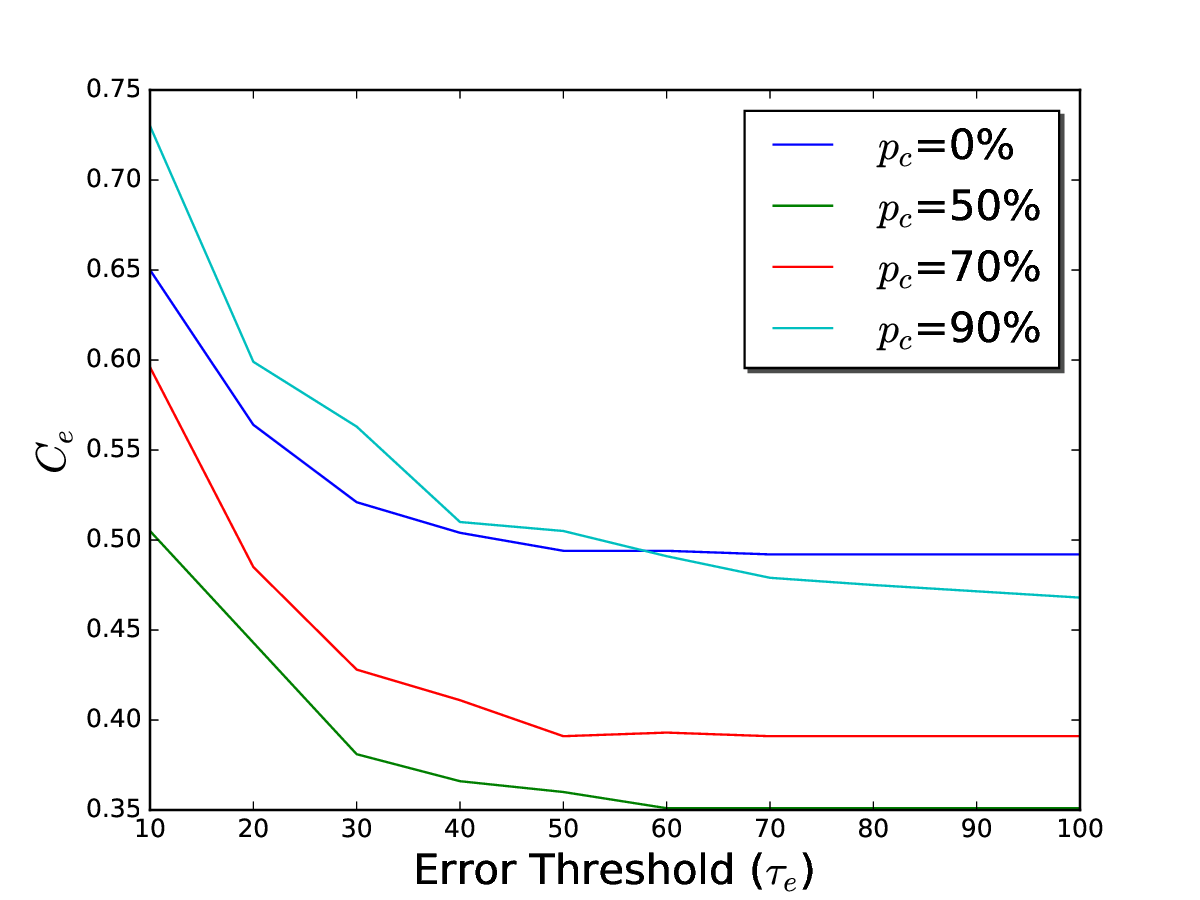}}
\caption{Uncertainty Evaluation w.r.t. Error.\label{fig:error-coverage}}
\end{figure*}
We compute $C_{e}$ at varying error threshold, $10 \leq \tau_e \leq 130$ and fixing uncertainty threshold $\tau_u$ as $0.2$. The results are shown in Figure \ref{fig:error-coverage}.
At lower $\tau_e$, higher $C_e$ indicates that higher fraction of correct predictions are confident in nature.

\begin{figure*}[h]
\subfigure[FD001 Dataset]{\includegraphics[trim={0cm 0cm 0cm 0cm},clip,width=0.4\textwidth]{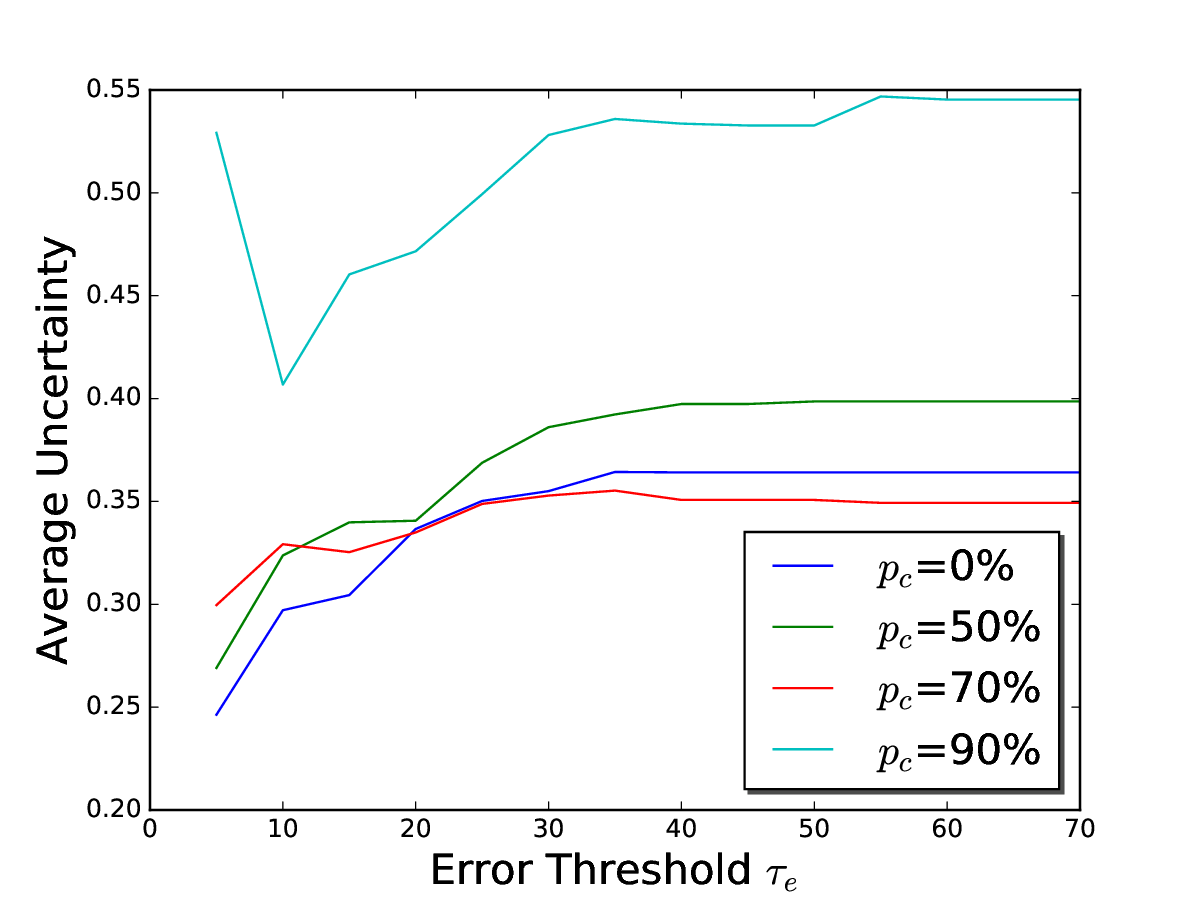}}\hspace{0.1\textwidth}
\subfigure[FD004 Dataset]{\includegraphics[trim={0cm 0cm 0cm 0cm},clip,width=0.4\textwidth]{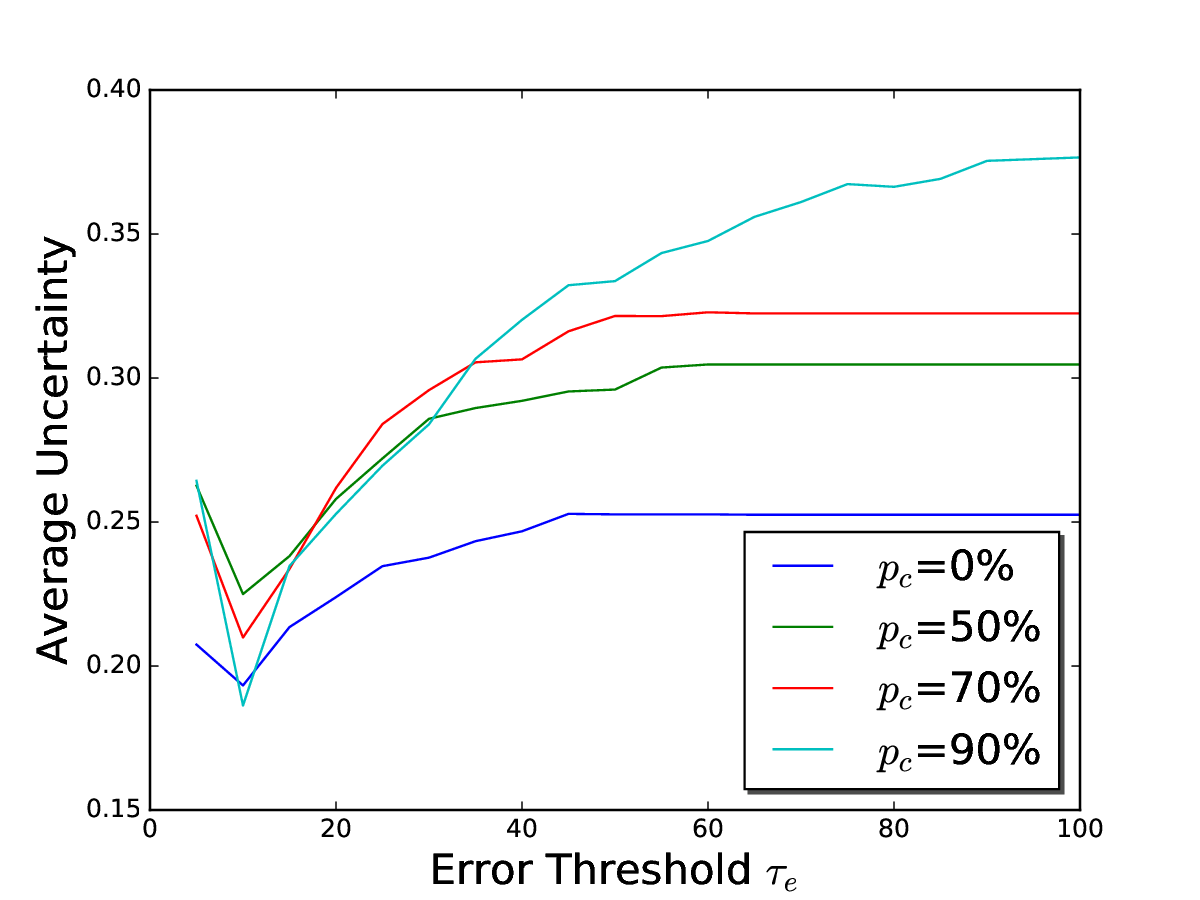}}
\caption{Average Uncertainty with varying $\tau_e$ indicating low uncertainty values when error in RUL estimates is low.\label{fig:sigma-error-plots}}
\end{figure*}

\textit{Relationship Between Error and Uncertainty}: RUL estimates with lower error values are expected to be certain in nature. For evaluating the uncertainty quantification metric from this aspect, we plot the relationship between error and average uncertainty, shown in Figure \ref{fig:sigma-error-plots}.
Average uncertainty value at given $\tau_e$ is computed by considering the test instances with RUL estimate error $\leq \tau_e$ and we then average out the uncertainty values corresponding to these filtered out test instances. 
We observe that at lower error thresholds, the computed average uncertainty is also low, indicating the preciseness of the RUL estimation model. Further, increase in average uncertainty with increase in error threshold indicates  reliable behavior of the RUL estimation model.

\end{document}